\documentclass[lettersize,journal]{IEEEtran}
\usepackage{amsmath,amsfonts}
\usepackage{algorithmic}
\usepackage{algorithm}
\usepackage{array}
\usepackage{textcomp}
\usepackage{stfloats}
\usepackage{url}
\usepackage{verbatim}
\usepackage{graphicx}
\usepackage{cite}

\usepackage{booktabs}
\usepackage{makecell}
\usepackage{subfigure}
\usepackage{multirow}
\usepackage{wasysym}
\usepackage{bm}
\usepackage{bbm}
\usepackage[table]{xcolor}

\begin{document}

\title{Image-Text Retrieval with Binary and Continuous Label Supervision}

\author{Zheng Li, Caili Guo,~\IEEEmembership{Senior Member,~IEEE}, Zerun Feng, Jenq-Neng Hwang,~\IEEEmembership{Fellow,~IEEE}, Ying Jin, Yufeng Zhang
\thanks{Caili Guo, Zheng Li and Zerun Feng are with School of Information and Communication Engineering, Beijing University of Posts and Telecommunications, Beijing 100876, China (e-mail: guocaili@bupt.edu.cn; lizhengzachary@bupt.edu.cn; fengzerun@bupt.edu.cn).}
\thanks{Jenq-Neng Hwang and Ying Jin are with the Department of Electrical Engineering, University of Washington, Seattle, WA 98105 USA (e-mail: hwang@uw.edu; jinying@uw.edu).}
\thanks{Yufeng Zhang is with China Telecom Dict Application Capability Center, Beijing, China (e-mail: zyf68@chinatelecom.cn).}
}



\maketitle

\begin{abstract}
Most image-text retrieval work adopts binary labels indicating whether a pair of image and text matches or not. 
Such a binary indicator covers only a limited subset of image-text semantic relations, which is insufficient to represent relevance degrees between images and texts described by continuous labels such as image captions.
The visual-semantic embedding space obtained by learning binary labels is incoherent and cannot fully characterize the relevance degrees.
In addition to the use of binary labels, this paper further incorporates continuous pseudo labels (generally approximated by text similarity between captions) to indicate the relevance degrees. 
To learn a coherent embedding space, we propose an image-text retrieval framework with \textit{Binary and Continuous Label Supervision} (\textit{BCLS}), where binary labels are used to guide the retrieval model to learn limited binary correlations, and continuous labels are complementary to the learning of image-text semantic relations.
For the learning of binary labels, we improve the common \textit{Triplet ranking loss with Soft Negative mining} (\textit{Triplet-SN}) to improve convergence.
For the learning of continuous labels, we design \textit{Kendall ranking loss} inspired by \textit{Kendall rank correlation coefficient} (\textit{Kendall} $ \tau $), which improves the correlation between the similarity scores predicted by the retrieval model and the continuous labels.
To mitigate the noise introduced by the continuous pseudo labels, we further design \textit{Sliding Window sampling and Hard Sample mining strategy} (\textit{SW-HS}) to alleviate the impact of noise and reduce the complexity of our framework to the same order of magnitude as the triplet ranking loss.
Extensive experiments on two image-text retrieval benchmarks demonstrate that our method can improve the performance of state-of-the-art image-text retrieval models.
We conduct an objective and fair comparison of existing retrieval methods with continuous label supervision based on the ECCV Caption dataset, which provides semantic associations for more image-text pairs.
The experimental results further demonstrate that our method can better learn continuous semantic relations.
\end{abstract}


\section{Introduction} \label{Introduction}
\IEEEPARstart{I}{mage-text} retrieval is formulated as retrieving relevant samples across the different image and text modalities \cite{faghri2018vse++, lee2018stacked, li2019visual, fu2019rich, zhang2022unified, wang2020pfan++, liu2022self, zhang2022negative, zhang2022show, cheng2022vista, cheng2022cross, li2022image}. 
In the case of image-to-text retrieval, given a query image, the goal is to find the most relevant caption from the text gallery. 
On the other hand, text-to-image retrieval starts with a query text, and the goal is to find the most relevant image from the image gallery. 
Compared with unimodal image retrieval, image-text retrieval is more challenging due to the heterogeneous gap between image and text. 
A dominant approach to deal with the above challenge is to learn a shared visual-semantic embedding space, where the distance between the embedding vectors of related image and text is minimized.

\begin{figure}[t]
	\centering
	\includegraphics[width=\linewidth]{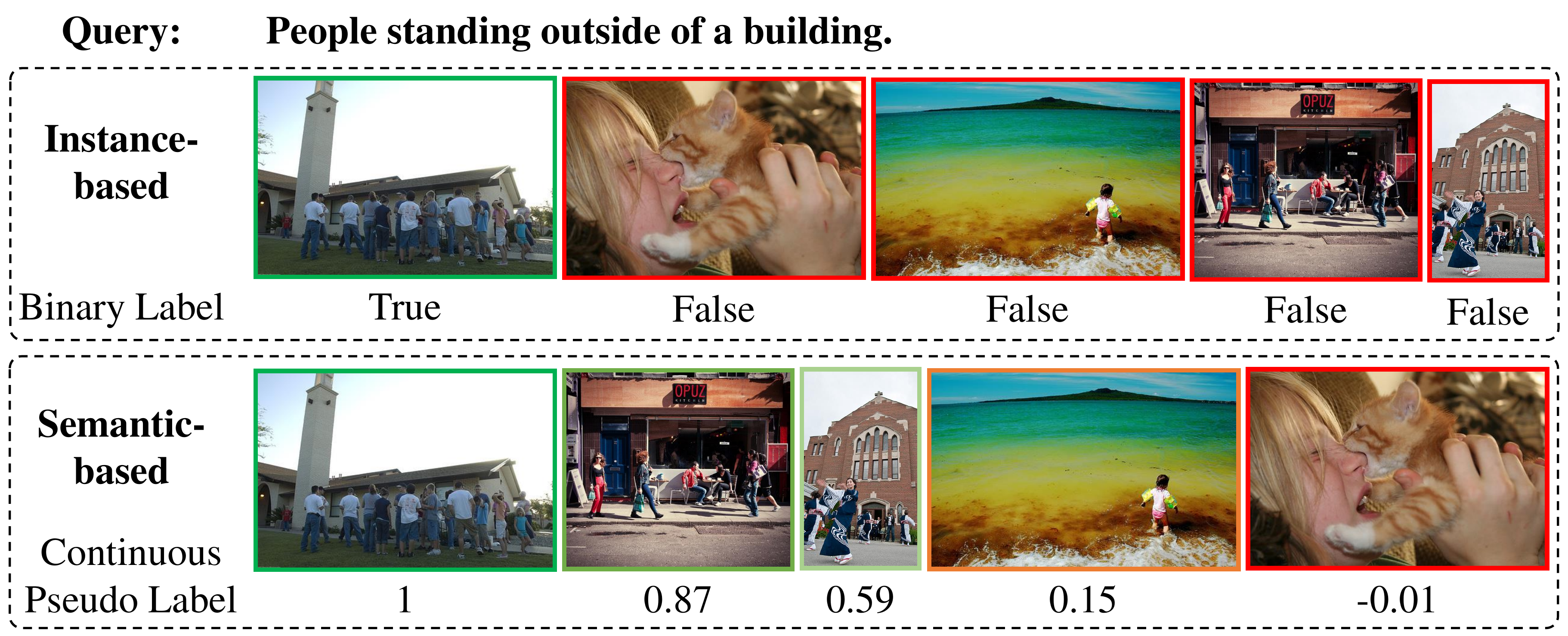}
	\caption{Instance-based retrieval adopts a binary label indicating whether a pair of image and text match or not.
	However, the relevance degrees between the samples marked as False and the query are different, and the binary label cannot reflect this relevance. 
	In semantic-based retrieval, continuous labels are used to model the relevance degrees between queries and candidates.
	The continuous pseudo labels in the figure are approximated by calculating text similarity using Sentence-BERT \cite{reimers2019sentence}.
	This allows multiple candidates to be considered relevant to a query and provides a way of ranking candidates from most to least similar.}
	\label{example}
\end{figure}
According to the assumption of the relevance between query and candidate, image-text retrieval can be mainly divided into two categories, instance-based and semantic-based, as shown in \tablename~\ref{classification}.
Most image-text retrieval work \cite{faghri2018vse++, lee2018stacked, li2019visual, fu2019rich, zhang2022unified, wang2020pfan++} focuses on instance-based retrieval. 
As shown in \figurename~\ref{example}, instance-based retrieval adopts a binary label indicating whether a pair of image and text match or not. 
Widely used image-text retrieval datasets such as Flickr30K \cite{young2014image} and MS-COCO \cite{lin2014microsoft} provide manually annotated binary labels.
Such a binary indicator covers only a limited subset of image-text semantic relations, which is insufficient to represent relevance degrees between images and texts described by continuous labels such as image captions. 
With binary label supervision, a query sentence is relevant to only one image. 
However, in fact, there may be multiple candidates related to the query that are directly arbitrarily classified as irrelevant. 
It's inconsistent with the user experience of the retrieval system. 
In practice, we hope that the retrieval system can return multiple results related to the query and rank them by relevance degrees since humans' judgment of relevance degree is not simply a binary relevance and irrelevance. 
Loss functions designed based on the above assumptions, such as the most widely used Triplet ranking loss with Hard Negative mining (Triplet-HN) \cite{faghri2018vse++}, cannot guide the model to learn a coherent visual-semantic embedding space.
Moreover, Triplet-HN only optimizes hardest negative samples, which will make the model training slow to converge and make optimization difficult.

\begin{table}[t]
	
	\caption{Image-text retrieval classification.}
	\begin{center}
		\begin{tabular}{p{0.19\linewidth}p{0.33\linewidth}p{0.33\linewidth}}
			\toprule[1pt]
			~ & \textbf{Instance-based} & \textbf{Semantic-based}\\
			\hline
			\specialrule{0em}{2pt}{0pt}
			Label & Binary label & Continuous pseudo label \\
			\hline
			\specialrule{0em}{2pt}{0pt}
			Label source & Human annotation & 
			Text similarity approximation \\
			\hline
			\specialrule{0em}{2pt}{0pt}
			Advantages & Binary labels are accurate and do not introduce training noise & Continuous pseudo labels can represent continuous correlations between images and texts \\
			\hline
			\specialrule{0em}{2pt}{0pt}
			Disadvantages & Binary labels cannot adequately represent the correlation between images and texts & Continuous pseudo labels are approximated, not completely accurate, and will introduce training noise \\
			\hline
			\specialrule{0em}{2pt}{0pt}
			Optimization objective & Triplet-HN \cite{faghri2018vse++} & Ladder loss \cite{zhou2020ladder, wang2021adaptive}, SAM loss \cite{biten2021image} \\
			\toprule[1pt]
		\end{tabular}
	\end{center}
	\label{classification}
\end{table}

In semantic-based retrieval, continuous labels are used to indicate the relevance degrees between queries and candidates, as \figurename~\ref{example} shows. 
It allows multiple candidates to be considered relevant to a query and provides a way of ranking candidates from most to least similar \cite{wray2021semantic}.
\tablename~\ref{classification} compares the advantages and disadvantages of the two types of retrieval methods.
Compared with instance-based retrieval, semantic-based retrieval is more in line with the actual user experience.
Although semantic-based retrieval is more preferred, it is difficult to obtain continuous labels. 
The ideal ground truth for the continuous label is human annotation, but it is infeasible to annotate an image-text pairwise relevance degree dataset.
Recently, some studies \cite{wray2021semantic,biten2021image, zhou2020ladder} present that the semantic similarities between texts can be used to approximate the relevance degrees between images and texts as the continuous pseudo labels. 
Wray \textit{et al.} \cite{wray2021semantic} propose several proxies to estimate relevance degrees. 
Zhou \textit{et al.} \cite{zhou2020ladder, wang2021adaptive} propose to measure the relevance degrees by BERT \cite{devlin2019bert} and design a ladder loss to learn a coherent embedding space. 
Biten \textit{et al.} \cite{biten2021image} use image captioning evaluation metrics to approximate the relevance degree, and design a Semantic Adaptive Margin (SAM) loss for semantic-based retrieval. 
Additionally, these studies also propose to use normalized Discounted Cumulative Gain (nDCG) \cite{wray2021semantic}, Coherent Score \cite{zhou2020ladder, wang2021adaptive} and Normalized Cumulative Semantic Score \cite{biten2021image} to evaluate the performance of semantic-based retrieval methods.
Compared with binary labels, continuous pseudo labels can represent the continuous correlation between images and texts. 
But since pseudo labels are calculated approximately, they are not completely accurate, which will introduce noise into training.

Existing semantic-based retrieval work has made great progress, but there are still the following problems:
\begin{itemize}
	\item Existing semantic-based retrieval methods achieves the learning of continuous pseudo labels by hierarchically embedding samples with different relevance degrees \cite{zhou2020ladder, wang2021adaptive}, or adjusting the margin of triplet loss according to pseudo labels \cite{biten2021image}. 
	These methods require manual selection of appropriate hyper-parameters and cannot be flexibly applied to different data and retrieval models.
	\item Continuous pseudo labels approximated by text similarity are not completely accurate, which will introduce noise into model training.
	Existing methods ignore the negative effects of inaccurate pseudo labels.
	\item Existing evaluation metrics of semantic-based retrieval can only reflect the fit of the retrieval model to inaccurate pseudo labels, which cannot objectively reflect the retrieval performance.
\end{itemize}

Using binary or continuous labels alone has its own shortcomings. 
This paper proposes an image-text retrieval framework with \textit{Binary and Continuous Label Supervision} (\textit{BCLS}) to learn a coherent visual-semantic embedding space.
For the learning of binary labels, we improve the common \textit{Triplet ranking loss with Soft Negative mining} (\textit{Triplet-SN}) to improve convergence.
For the learning of continuous labels, we design \textit{Kendall ranking loss} inspired by \textit{Kendall rank correlation coefficient} (\textit{Kendall} $ \tau $).
In statistics, \textit{Kendall} $ \tau $ is a statistic used to measure the ordinal association between two measured quantities. 
This loss function improves the correlation between the similarity scores predicted by the retrieval model and the continuous pseudo labels by optimizing the discordant pairs in the ranking results, which can be flexibly applied to various data and retrieval models without complex parameter settings.
For the problem of pseudo labels introducing training noise, we further design \textit{Sliding Window sampling and Hard Sample mining strategy} (\textit{SW-HS}) to alleviate the impact of noise and reduce the complexity of our framework to the same order of magnitude as the common triplet ranking loss.
For the evaluation problem of semantic-based retrieval, we conducted an objective and fair comparison of existing semantic-based retrieval methods with the help of the Extended COCO Validation (ECCV) Caption dataset \cite{chun2022eccv}.
This dataset leverages machine and human annotations to provide semantic associations for more image-text pairs.
The major contributions of this paper are summarized as follows:
\begin{itemize}
	\item A novel image-text retrieval framework with \textit{Binary and Continuous Label Supervision} (\textit{BCLS}) is proposed to guide retrieval models to learn a coherent visual-semantic embedding space.
	The framework combines the advantages of binary and continuous labels and makes targeted improvements for the problems existing in the two types of label learning.
	\item A \textit{Sliding Window sampling and Hard Sample mining strategy} (\textit{SW-HS}) is designed to mitigate the negative effects of continuous pseudo label inaccuracy and reduce the complexity of framework with \textit{BCLS} to the same order of magnitude as the common triplet ranking loss.
	\item To address the shortcomings of performance evaluation for semantic-based retrieval, we conduct an objective and fair comparison of existing semantic-based retrieval methods with the help of the ECCV Caption dataset, which provides semantic associations for more image-text pairs.
	The experimental results demonstrate that our method can better learn continuous semantic relations.
\end{itemize}

\section{Related Work}
\subsection{Instance-based Image-Text Retrieval}
Image-text retrieval task, either image-to-text or text-to-image, is formulated as retrieving relevant samples across the different image and text modalities \cite{faghri2018vse++, lee2018stacked, li2019visual, fu2019rich, zhang2022unified, wang2020pfan++, liu2022self, zhang2022negative, zhang2022show, cheng2022vista, cheng2022cross, li2022image}. 
According to the assumption of the relevance between query and candidate, image-text retrieval can be mainly divided into two categories, instance-based and semantic-based.
Most image-text retrieval studies \cite{faghri2018vse++, lee2018stacked, li2019visual, fu2019rich, zhang2022unified, wang2020pfan++} focus on instance-based retrieval. 
A variety of methods have been devoted to learning modality invariant features. 
More specifically, Wang \textit{et al.} \cite{wang2020pfan++} propose a position focused attention network to investigate the relation between the visual and the textual views for image-text retrieval. 
In recent years, multi-modal pre-training models \cite{lu2019vilbert, chen2020uniter, lu202012, li2020oscar, jia2021scaling, radford2021learning, li2021align, zhang2021vinvl} have been intensively explored to bridge image and text.
The paradigm of vision-language pre-training is to design pre-training tasks on large-scale vision-language data for pre-training and then finetune the model on specific downstream tasks.
The above methods learn advanced encoding networks to generate richer semantic representations for different modalities.
The framework with \textit{BCLS} proposed in this paper is independent of image and text feature representation and similarity calculation. 
It can be plug-and-play applied to existing instance-based retrieval models.

In addition to the work on the feature representation and similarity calculation of images and text, a variety of deep metric learning methods have been proposed in instance-based image-text retrieval \cite{faghri2018vse++, wei2020universal, chen2020interclass, chen2020adaptive}. 
A hinge-based triplet loss is widely employed as an objective to enforce aligned pairs to have a higher similarity score than misaligned pairs by a margin \cite{frome2013devise}. 
Faghri \textit{et al.} \cite{faghri2018vse++} incorporate hard negatives in the ranking loss function, which yields significant gains in retrieval performance. 
There are several studies \cite{wei2020universal, chen2020interclass, wei2021universal} that propose weighting metric learning frameworks for image-text retrieval, which can further improve retrieval performance.
These loss functions for instance-based image-text retrieval adopt binary labels to indicate whether a pair of image and text match or not, which is not sufficient to represent the relevance degree between image and text.
Using a binary label based loss function to train a model will destroy the coherence of the visual-semantic embedding space, making it difficult for the model to learn continuous semantic relations.

\subsection{Semantic-based Image-Text Retrieval}
While most work focuses on instance-based retrieval, a few studies have explored semantic-based retrieval.  
Some studies propose that the semantic similarity between captions can be used to approximate the relevance degree between image and text \cite{wray2021semantic, zhou2020ladder}. 
Wray \textit{et al.} \cite{wray2021semantic} propose several proxies to estimate relevance degrees. 
Biten \textit{et al.} \cite{biten2021image} use image captioning evaluation metrics, \textit{i.e.}, Consensus-based Image Description Evaluation (CIDEr) \cite{vedantam2015cider} and Semantic Propositional Image Caption Evaluation (SPICE) \cite{anderson2016spice}, to approximate the relevance degree, and design a semantic adaptive margin (SAM) loss for semantic-based retrieval. 
SAM loss is a variant of triplet loss, where candidates are pushed away from the query by semantic adaptive margins in the embedding space.
The adjustment range of the margin in SAM loss will be affected by the dataset, retrieval model, and pseudo label calculation method. 
It needs to be carefully adjusted manually and cannot be flexibly applied to different data and models. 
Zhou \textit{et al.} \cite{zhou2020ladder, wang2021adaptive} propose to measure the relevance degrees by BERT \cite{devlin2019bert} and design a ladder loss to learn a coherent embedding space. 
In the ladder loss, the relevance degrees are artificially divided into several levels, and a large number of hyper-parameters are introduced.
Moreover, pseudo labels approximated by text similarity are not completely accurate, and existing methods ignore the negative effects of inaccurate pseudo labels.
The \textit{Kendall ranking loss} proposed in this paper will solve the problems existing in the current semantic-based retrieval methods.

In addition to methodological problems, performance evaluation of semantic-based retrieval methods also has shortcomings. 
Existing evaluation metrics of semantic-based retrieval can only reflect the fit of the retrieval model to inaccurate pseudo labels, which cannot objectively reflect the retrieval performance.
This paper will make up for the existing shortcomings in the performance evaluation of semantic-based retrieval.

\subsection{Deep Metric Learning}
The main work of this paper belongs to the field of deep metric learning. 
Deep metric learning aims to construct an embedding space to reflect the semantic distances among instances. 
It has many other applications such as face recognition \cite{schroff2015facenet} and image retrieval \cite{oh2016deep}. 
Contrastive loss \cite{hadsell2006dimensionality} and triplet loss \cite{hoffer2015deep} are two representative pairwise approaches in deep metric learning. 
Unlike the contrastive loss, which aims to push misaligned pairs apart by a fixed margin as well as to pull aligned pairs as close as possible. 
Triplet loss only aims to force the similarity of a positive pair to be higher than that of a negative one by a margin and enjoys more flexibility. 
Unsatisfied with potential slow convergence and unstable performance, recent work have proposed several variants. 
N-pair loss \cite{sohn2016improved} employed multiple negatives for each positive sample. 
However, the above methods are all applied to unimodal image retrieval, where relevance degrees of the instances can be clearly defined as a binary variable.
These loss functions can only guide the model to map images of the same class to relatively close locations and images of different classes to distant locations, and cannot be used to learn continuous semantic relationships between images and text.

Recently, some methods \cite{cakir2019deep, brown2020smooth} for directly optimizing evaluation metrics such as average precision (AP) have been proposed. Cakir \textit{et al.} \cite{cakir2019deep} propose FastAP to optimize AP using a soft histogram binning technique. Brown \textit{et al.} \cite{brown2020smooth}, on the other hand, optimize a smoothed approximation of AP, called Smooth-AP. Direct optimization of evaluation metrics looks at more samples from the retrieval set and has been proven to improve training efficiency and performance \cite{brown2020smooth}.
Inspired by the above work, this paper designs the \textit{Kendall ranking loss} to optimize the \textit{Kendall} $ \tau $ between the ranking scores predicted by the retrieval model and the continuous pseudo labels.

\section{Image-Text Retrieval with Binary and Continuous Label Supervision}
\begin{figure*}[!t]
	\centering
	\includegraphics[width=\linewidth]{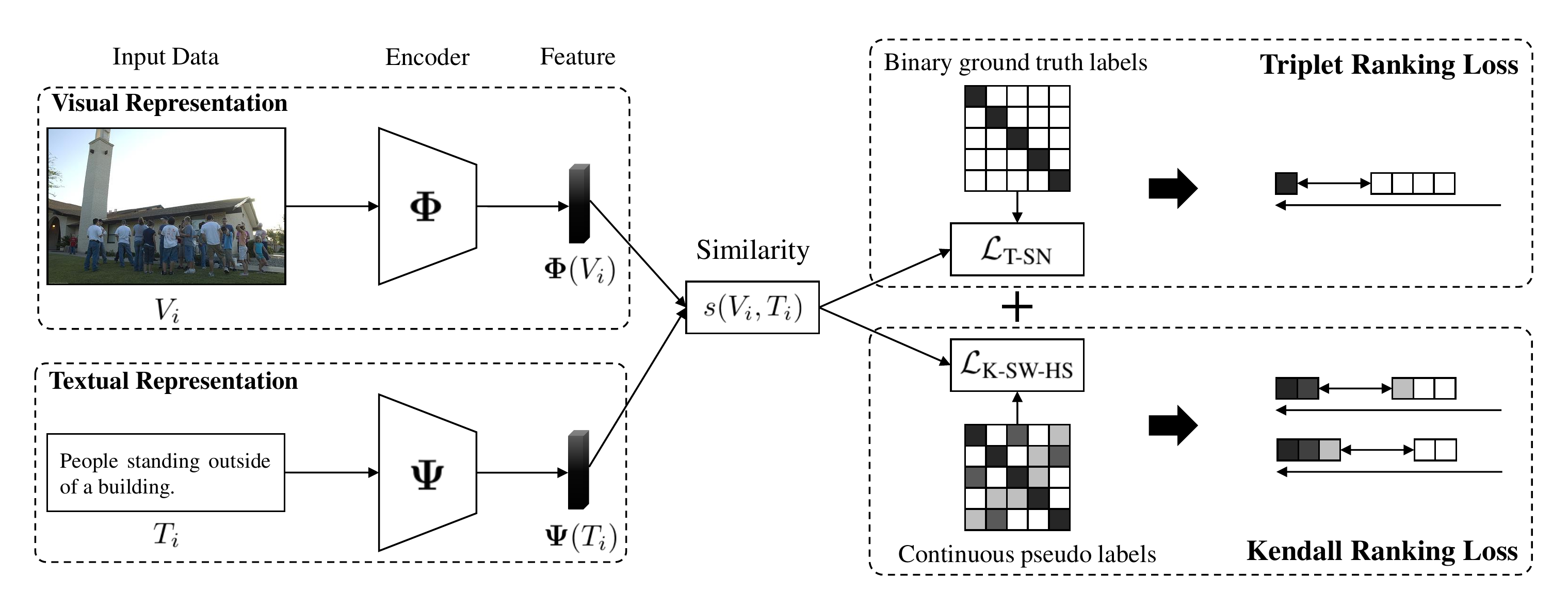}
	\caption{Image-text retrieval framework with \textit{Binary and Continuous Label Supervision} (\textit{BCLS}). 
	The framework is divided into two branches: binary label and continuous pseudo label. 
	The binary label branch uses triplet ranking loss as the optimization objective. The continuous pseudo label branch uses \textit{Kendall ranking loss} as the optimization objective.
	The color of the small square representing the label in the figure represents the size of the label. 
	The darker the color, the higher the relevance degree between the query and the candidate.}
	\label{framework}
\end{figure*}
\subsection{Preliminaries} \label{Preliminaries}
In this section, we first introduce the related background of image-text retrieval in detail.
Image-text retrieval task, either image-to-text or text-to-image, is formulated as retrieving relevant samples across the different image and text modalities.
Formally, given a set of images $ \bm{V} $ and a corresponding set of captions $ \bm{T} $. Let $ V_{i} $ be a image, $ T_{i} $ be a sentence. 
In the case of image-to-text retrieval, given a query image $ V_{i} $, the goal is to find the most relevant caption $ T_{i} $ from the text gallery. 
On the other hand, text-to-image retrieval starts with a query text $ T_{i} $, and the goal is to find the most relevant image $ V_{i} $ from the image gallery. 
Previous work for image-text retrieval focuses on building a shared visual-semantic embedding space that contains both the image and sentence. 
The core idea behind these methods is that there exists a mapping function, $ s(V_{i}, T_{i}; \bm{W}) = \bm{\Phi}(V_{i})^{\top} \bm{W} \bm{\Psi}(T_{i}) $  to measure the similarity score between the image features $ \bm{\Phi}(V_{i}) $ and the text features $ \bm{\Psi}(T_{i}) $, where $ \bm{W} $ is the parameter of $ s(\cdotp, \cdotp) $.

Most image-text retrieval work focuses on instance-based retrieval, which defines a binary label $ y(V_{i}, T_{j}) $ between an image $ V_{i} $ and a caption $ T_{j} $. 
$ y(V_{i}, T_{j}) = 1 $, when $ i = j $ and $ y(V_{i}, T_{j}) = 0 $, when $ i \neq j $.
Widely used image-text retrieval datasets such as Flickr30K \cite{young2014image} and MS-COCO \cite{lin2014microsoft} provide manually annotated binary labels.
Only one related image per caption and the captions of different images are assumed to be irrelevant. 
However, it is common for multiple similar images and captions to appear in large datasets. 
Besides, loss functions designed based on the above assumptions, such as the most widely used Triplet ranking loss with Hard Negative mining (Triplet-HN) \cite{faghri2018vse++}, cannot guide the model to learn a coherent visual-semantic embedding space.

Semantic-based image-text retrieval defines a continuous label, $ r(V_{i}, T_{j}) \rightarrow [-1, 1] $, that captures the relevance degree between any image-caption pair. 
When $ r(V_{i}, T_{j}) = 1 $, $ V_{i} $ and $ T_{j} $ are regarded as completely relevant. The more relevant $ V_{i} $ and $ T_{j} $ is, the closer $ r(V_{i}, T_{j}) $ is to $ 1 $, and the less relevant, the closer $ r(V_{i}, T_{j}) $ to $ -1 $. 
Multiple captions can have the same similarity to an image, and vice versa for multiple images to the same caption. 
In addition, the continuous label can model varying levels of similarity. 
If $ r(V_{i}, T_{j}) > r(V_{i}, T_{k}) $, $ T_{j} $ is a more relevant caption to the image $ V_{i} $ than $ T_{k} $. 
If $ r(V_{i}, T_{j}) = r(V_{i}, T_{k}) $, $ T_{j} $ and $ T_{k} $ are considered equally relevant to $ V_{i} $ and ranking them in any order should not be penalized by the evaluation metric. 
Compared with instance-based retrieval, semantic-based retrieval is more in line with the actual user experience.

\begin{table}[t]
	\caption{Pearson Correlation Coefficient between the approximate relevance degrees and human judgments on CxC dataset.}
	\begin{center}
		\begin{tabular}{ccccccccccccc}
			\toprule[1pt]
			~ & CLIP & CIDEr & METEOR & SPICE & S-BERT \\
			\hline
			\specialrule{0em}{2pt}{0pt}
			Pearson & 0.373 & 0.738 & 0.757 & 0.728 & 0.877 \\
			\bottomrule[1pt]
		\end{tabular}
	\end{center}
\label{pearson}
\end{table}
Although semantic-based retrieval is more advantageous, it is difficult to obtain continuous relevance degree labels. 
We learn from the proxy measures for relevance degree in \cite{zhou2020ladder}. 
For image-text retrieval tasks, we are more concerned about the semantic relevance degree between samples. 
NLP methods are more suitable for measuring semantic relevance degree than visual methods. 
Thanks to the rapid development of NLP, like BERT \cite{devlin2019bert}, which has nearly reached human performance on the sentence similarity task, we cast the image-text relevance degree measure problem as a text relevance degree measure problem. 
For an image $ V_{i} $, the relevance degree of its corresponding caption $ T_{i} $ is supposed to be $ 1 $, and it is regarded as a reference when measuring the relevance degrees between $ V_{i} $ and other captions.
Following Zhou \textit{et al.} \cite{zhou2020ladder}, we employ Sentence-BERT (S-BERT) \cite{reimers2019sentence} to compute inter-text similarity as continuous pseudo labels for semantic-based retrieval.
The pseudo label $ r(V_{i}, T_{j}) $ between $ V_{i} $ and $ T_{j} $ can be approximated as the normalized cosine similarity $ s(T_{i}, T_{j}) $ between $ T_{i} $ and $ T_{j} $.

To verify the accuracy of the continuous pseudo labels we generate, we choose to use the Crisscrossed Captions (CxC) dataset \cite{parekh2021crisscrossed} to compare the accuracy of several methods for generating annotations.
CxC dataset is an extension of MS-COCO \cite{lin2014microsoft} that comprises human judgments  on the degree of semantic similarity between captions and images. 
These judgments  are collected not only for the predefined ground truth pairs but also for other pairs. 
CxC dataset contains human semantic similarity judgments for 267,095 intra- and inter-modality pairs. 
The dataset contains 44,844 caption-image pairs, of which 25,000 pairs are originally labeled as relevant in the MS-COCO. 
We can compare the accuracy of methods that generate annotations by comparing how close the generated annotations are to human annotations.
We compute the Pearson correlation coefficients between several pseudo label generation methods and human judgments.
Comparison methods include some image captioning evaluation metrics METEOR \cite{banerjee2005meteor}, CIDEr \cite{vedantam2015cider}, SPICE \cite{anderson2016spice} and visual language pre-trained model CLIP \cite{radford2021learning}.
As shown in \tablename~\ref{pearson}, the pseudo labels computed by S-BERT are closer to human judgment and much better than other approximation methods and the pre-trained model CLIP.
Compared with binary labels, continuous pseudo labels can represent the continuous correlation between images and texts. 
But since pseudo labels are calculated approximately, they are not completely accurate, which will introduce noise into training.

\subsection{Image-Text Retrieval Framework with Binary and Continuous Label Supervision}
Binary labels can guide the model to learn accurate binary correlations between images and texts. 
Continuous labels are complementary to the learning of image-text semantic relations.
Using binary or continuous labels alone has its own shortcomings. 
Therefore, we not only use the original binary labels of the dataset but also use the continuous pseudo labels generated by NLP methods as supervision for model training.
This paper proposes a image-text retrieval framework with \textit{Binary and Continuous Label Supervision} (\textit{BCLS}), as shown in \figurename~\ref{framework}. The framework is divided into two branches: binary label and continuous pseudo label.

\subsubsection{Binary Label Branch}
The binary label branch uses triplet ranking loss as the optimization objective. 
The classic triplet ranking loss without sampling can be written as:
\begin{equation} \label{triplet}
	\begin{aligned}
		\mathcal{L}_{\text{T}}
		= & \sum_{i=1}^{B} 
		\sum_{j=1, i \neq j}^{B} 
		\left[ 
		s(V_{i}, T_{j}) - s(V_{i}, T_{i}) + m 
		\right]_{+} \\
		+ & \sum_{i=1}^{B} 
		\sum_{j=1, i \neq j}^{B} 
		\left[ 
		s(V_{j}, T_{i}) - s(V_{i}, T_{i}) 
		+ m
		\right]_{+}, 
	\end{aligned}
\end{equation}
where $ [x]_{+} = \max (x, 0) $, $ B $ is the batch size and $ m $ is a margin for better similarity separation.
$ \mathcal{L}_{\text{T}} $ guides the model to learn an embedding space where the similarity of positive sample pairs $ s(V_{i}, T_{i}) $ is greater than the similarity of negative sample pairs $ s(V_{i}, T_{j}) $ plus a margin $ m $.
Triplet ranking loss tends to treat the relevance between queries and candidates in a bipolar way. 
Therefore, triplet loss cannot learn the continuous relevance degree between samples.

\subsubsection{Continuous Label Branch}
For the branch supervised by continuous pseudo labels, we design a loss function based on the \textit{Kendall rank correlation coefficient} (\textit{Kendall} $ \tau $) as the optimization objective.
\textit{Kendall} $ \tau $ is a suitable evaluation metric for evaluating retrieval tasks with continuous labels. 
In statistics, \textit{Kendall} $ \tau $ is a statistic used to measure the ordinal association between two measured quantities. \textit{Kendall} $ \tau $ is defined as:
\begin{equation}\label{kendall}
	\tau = \frac{C-D}{N(N-1)/2},
\end{equation}
where $ C $ is the number of concordant pairs, $ D $ is the number of discordant pairs, $ N(N-1)/2 $ is the number of ways to choose two items from $ N $ items.

From Eq.~\eqref{kendall}, it can be seen that optimizing the discordant pairs in the ranking results can improve the correlation between the similarity scores predicted by the retrieval model and the continuous pseudo labels.
Since \textit{Kendall} $ \tau $ is a statistic, which is not differentiable to the embedding vector of the sample, we need to transform \textit{Kendall} $ \tau $ into a function that can be derived from the embedding vector. 
In Eq.~\eqref{kendall}, $ C $ and $ D $ are the numbers of concordant and discordant pairs, respectively. 
We can transform the relationship between concordant and discordant pairs into an inequality relationship about sample similarity:
\begin{equation}
	\begin{aligned}
		& s(V_{i}, T_{j}) > s(V_{i}, T_{k}),  r(V_{i}, T_{j}) > r(V_{i}, T_{k}), \text{concordant},\\
		& s(V_{i}, T_{j}) < s(V_{i}, T_{k}),  r(V_{i}, T_{j}) > r(V_{i}, T_{k}), \text{discordant}.
	\end{aligned}
\end{equation}
That is to say, under ideal circumstances, the inequality relationship of sample similarity should be consistent with the inequality relationship of continuous pseudo labels. 
To simplify the representation, we abbreviate $ r(V_{i}, T_{j}) $ as $ r_{ij} $.
If we consider satisfying the above inequality relationship as the optimization goal, the \textit{Kendall ranking loss} can be written as:
\begin{equation}
	\begin{aligned}
		\mathcal{L}_{\text{K}}
		= & \sum_{i=1}^{B}
		\sum_{j=1}^{B}
		\sum_{k=1}^{B}
		\mathbbm{1}(r_{ij} > r_{ik})
		\cdot
		\left[ 
		s(V_{i}, T_{k}) - s(V_{i}, T_{j}) 
		\right]_{+} \\
		+ & \sum_{i=1}^{B}
		\sum_{j=1}^{B}
		\sum_{k=1}^{B}
		\mathbbm{1}(r_{ji} > r_{ki})
		\cdot
		\left[ 
		s(V_{k}, T_{i}) - s(V_{j}, T_{i}) 
		\right]_{+},
	\end{aligned}
\end{equation}
where
\begin{equation}
	\mathbbm{1} \{ x \} =
	\begin{cases}
		1 ,& \text{if $ x $ is true}, \\
		0 ,& \text{otherwise},
	\end{cases}
\end{equation}
and $ [x]_{+} = \max (x, 0) $. 

We can determine that $ \mathcal{L}_{\text{K}} $ is a differentiable function of the feature vectors, thus proving that it can be solved using gradient descent.
We consider the neural networks $ \bm{\Phi}(\cdot) $ and $ \bm{\Psi}(\cdot) $ embed the image and text on a unit hypersphere. 
We use $ (\bm{v}_{i}, \bm{t}_{i}) $ to simplify the representation of the normalized feature vectors. 
When embedded on a unit hypersphere, the cosine similarity is a convenient metric to measure the similarity between image-text pair $ s(V_{i}, T_{i}) = \bm{v}_{i}^{\top} \bm{t}_{i} $, and this similarity is bounded in the range $ [-1, 1] $. 

To simplify the representation, we only analyze the loss for image-to-text retrieval, which is symmetric to the loss from text-to-image retrieval.
\textit{Kendall ranking loss} for image-to-text retrieval can be written as:
\begin{equation}
	\mathcal{L}_{\text{K}}
	(\bm{v}_{i})
	= \sum_{j=1}^{B}
	\sum_{k=1}^{B}
	\mathbbm{1}(r_{ij} > r_{ik})
	\cdot
	\max \left( 
	\bm{v}_{i}^{\top} \bm{t}_{k} - \bm{v}_{i}^{\top} \bm{t}_{j} + m, 0
	\right).
\end{equation}
We derive the loss gradient with respect to the feature vectors $ \bm{v}_{i} $, $ \bm{t}_{j} $ and $ \bm{t}_{k} $:
\begin{equation}
	\begin{aligned}
		\frac{
			\partial 
			\mathcal{L}_{\text{K}}
			(\bm{v}_{i})}
		{\partial \bm{v}_{i}} 
		= & \sum_{j=1}^{B}
		\sum_{k=1}^{B}
		\mathbbm{1}(r_{ij} > r_{ik}) \\
		& \cdot 
		\mathbbm{1}
		(\bm{v}_{i}^{\top} \bm{t}_{k} - \bm{v}_{i}^{\top} \bm{t}_{j} + m > 0 )
		\cdot
		( \bm{t}_{k} - \bm{t}_{j} ),
	\end{aligned}
\end{equation}
\begin{equation}
	\frac{
		\partial 
		\mathcal{L}_{\text{K}}
		(\bm{v}_{i})}
	{\partial \bm{t}_{j}} 
	= - \mathbbm{1}(r_{ij} > r_{ik}) \cdot 
	\mathbbm{1}
	(\bm{v}_{i}^{\top} \bm{t}_{k} - \bm{v}_{i}^{\top} \bm{t}_{j} + m > 0 )
	\cdot
	\bm{v}_{i},
\end{equation}
\begin{equation}
	\frac{
		\partial 
		\mathcal{L}_{\text{K}}
		(\bm{v}_{i})}
	{\partial \bm{t}_{k}} 
	= \mathbbm{1}(r_{ij} > r_{ik}) \cdot 
	\mathbbm{1}
	(\bm{v}_{i}^{\top} \bm{t}_{k} - \bm{v}_{i}^{\top} \bm{t}_{j} + m > 0 )
	\cdot
	\bm{v}_{i}.
\end{equation}
$ \mathcal{L}_{\text{K}} (\bm{v}_{i}) $ is differentiable for all three feature vectors $ \bm{v}_{i} $, $ \bm{t}_{j} $ and $ \bm{t}_{k} $ associated with it.
Therefore, \textit{Kendall ranking loss} can be solved using gradient descent method.

$ \mathcal{L}_{\text{K}} $ is the basic form of \textit{Kendall ranking loss}, which filters out the discordant pairs through the indicator function $ \mathbbm{1} \{ x \} $. 
The objective of $ \mathcal{L}_{\text{K}} $ is to optimize all discordant pairs into concordant pairs, so as to improve the correlation between the similarity scores predicted by the retrieval model and the continuous pseudo labels.
$ \mathcal{L}_{\text{K}} $ realizes the learning of continuous pseudo labels by optimizing discordant pairs without introducing hyper-parameters,
which can be flexibly applied to various data and retrieval models.

During training, we add the loss functions of the two branches as the final optimization objective of our framework with \textit{BCLS}.
Binary labels can only guide the retrieval model to learn limited binary correlations, and continuous pseudo labels are complementary to the learning of image-text semantic correlations.
Optimizing both loss functions at the same time will not conflict.
If the sample pairs with $ r(\cdot, \cdot)=1 $ are regarded as positive pairs, and the sample pairs with $ r(\cdot, \cdot)<1 $ are regarded as negative pairs. 
Eq.~\eqref{triplet} can be written as:
\begin{equation}
	\begin{aligned}
		\mathcal{L}_{\text{T}}
		= & \sum_{i=1}^{B} 
		\sum_{j=1, r_{ij} < 1}^{B} 
		\left[ 
		s(V_{i}, T_{j}) - s(V_{i}, T_{i}) + m 
		\right]_{+} \\
		+ & \sum_{i=1}^{B} 
		\sum_{j=1, r_{ji} < 1}^{B} 
		\left[ 
		s(V_{j}, T_{i}) - s(V_{i}, T_{i}) 
		+ m
		\right]_{+}. 
	\end{aligned}
\end{equation}
When $ i=j $, $ \mathcal{L}_{\text{K}} (i=j) $ is the same as $ \mathcal{L}_{\text{T}} (m=0) $:
\begin{equation}
	\begin{aligned}
		\mathcal{L}_{\text{K}} (i=j)
		= & \sum_{i=1}^{B}
		\sum_{k=1}^{B}
		\mathbbm{1}(r_{ii} > r_{ik})
		\cdot
		\left[ 
		s(V_{i}, T_{k}) - s(V_{i}, T_{i}) 
		\right]_{+} \\
		+ & \sum_{i=1}^{B}
		\sum_{k=1}^{B}
		\mathbbm{1}(r_{ii} > r_{ki})
		\cdot
		\left[ 
		s(V_{k}, T_{i}) - s(V_{i}, T_{i}) 
		\right]_{+} \\
		= & \mathcal{L}_{\text{T}} (m=0).
	\end{aligned}
\end{equation}
$ \mathcal{L}_{\text{T}} (m=0) $ is a special form of the $ \mathcal{L}_{\text{K}} $ in the case of binary label supervision. 
The optimization goals of the two loss functions are consistent, and there will be no conflict in optimizing the two loss functions at the same time.

\subsection{Triplet Ranking Loss with Soft Negative Mining}
During the training process, there may be many samples in a batch that already meet the constraints of the loss function. 
These samples will no longer play a positive role in the optimization of the model, so it is important to mine hard samples. 
$ \mathcal{L}_{\text{T}} $ optimizes all negative sample pairs fairly, which will cost the model performance. 
Faghri \textit{et al.} \cite{faghri2018vse++} incorporate hard negatives in the ranking loss function, which yields significant gains in retrieval performance. 
Triplet ranking loss with Hard Negative mining (Triplet-HN) can be written as:
\begin{equation}
	\begin{aligned}
		\mathcal{L}_{\text{T-HN}}
		= & \sum_{i=1}^{B} 
		\left[ m - s(V_{i}, T_{i}) 
		+ s(V_{i}, \hat{T}_{i}) \right]_{+} \\
		+ & \sum_{i=1}^{B} 
		\left[ m - s(V_{i}, T_{i}) 
		+ s(\hat{V}_{i}, T_{i}) \right]_{+}, 
	\end{aligned}
\end{equation}
where
\begin{equation} \label{hn}
	\begin{aligned}
		\hat{T}_{i} 
		= & \arg \max _{j=1, r_{ij} < 1}^{B}
		s(V_{i}, T_{j}), \\
		\hat{V}_{i} 
		= & \arg \max _{j=1, r_{ji} < 1}^{B}
		s(V_{j}, T_{i}).
	\end{aligned}
\end{equation}
$ \mathcal{L}_{\text{T-HN}} $ is currently the most commonly used loss function in image-text retrieval tasks, and many state-of-the-art models \cite{faghri2018vse++, li2019visual, diao2021similarity} use this loss. 
However, $ \mathcal{L}_{\text{T-HN}} $ only mines one hardest negative sample in each batch, which will affect the convergence and increase the difficulty of model optimization. 
The framework with \textit{BCLS} needs to learn two types of labels at the same time, which further increases the training difficulty. 

To address the above issues, we propose \textit{Triplet ranking loss with Soft Negative mining} (\textit{Triplet-SN}):
\begin{equation}
	\begin{aligned}
		\mathcal{L}_{\text{T-SN}}
		= & \sum_{i=1}^{B} 
		\left[ 
		\tilde{s}(V_{i}) - s(V_{i}, T_{i})
		+ m
		\right]_{+} \\
		+ & \sum_{i=1}^{B} 
		\left[ 
		\tilde{s}(T_{i}) - s(V_{i}, T_{i})
		+ m
		\right]_{+}, 
	\end{aligned}
\end{equation}
where
\begin{equation}
	\begin{aligned}
		\tilde{s}(V_{i})
		= & \frac{1}{\gamma}
		\log \left(
		\sum_{j=1, r_{ij} < 1}^{B} 
		\exp{ 
			\left( 
			\gamma \cdot s(V_{i}, T_{j}) 
			\right)
		}
		\right), \\
		\tilde{s}(T_{i})
		= & \frac{1}{\gamma}
		\log \left(
		\sum_{j=1, r_{ji} < 1}^{B} 
		\exp{ 
			\left( 
			\gamma \cdot s(V_{j}, T_{i})
			\right)
		}
		\right).
	\end{aligned}
\end{equation}
We use the Log-Sum-Exp function to approximate the maximum function in Eq.~\eqref{hn}. 
$ \gamma $ is a scale factor to control the hardness of mining hard samples. 
When $ \gamma \rightarrow +\infty $, $ \mathcal{L}_{\text{T-SN}} $ is transformed into $ \mathcal{L}_{\text{T-HN}} $:
\begin{equation}
	\begin{aligned}
		\lim_{ \gamma \rightarrow +\infty}
		\tilde{s}(V_{i})
		= & \max _{j=1, r_{ij} < 1}^{B}
		s(V_{i}, T_{j}), \\
		\lim_{ \gamma \rightarrow +\infty} 
		\tilde{s}(T_{i})
		= & \max _{j=1, r_{ji} < 1}^{B}
		s(V_{j}, T_{i}),
	\end{aligned}
\end{equation}
$ \gamma \rightarrow +\infty $ represents that the loss function only mines the hardest negative samples in a batch.
When $ \gamma $ is a constant value, $ \mathcal{L}_{\text{T-SN}} $ will combine all negative samples to approximate the similarity $ \tilde{s}(\cdot) $ of a pair of hard negative samples. 
This design can improve the convergence of training and reduce the difficulty of model optimization.

\subsection{Kendall Ranking Loss with Sliding Window Sampling and Hard Sample Mining}
Although \textit{Kendall} $ \tau $ is transformed into a derivative function $ \mathcal{L}_{\text{K}} $, there is still problems with using $ \mathcal{L}_{\text{K}} $ as an optimization target. 
Since the continuous pseudo label $ r_{ij} $ is approximated by the similarity between the texts, it is not completely accurate.
Using inaccurate pseudo labels as supervision will introduce noise into model training.
For approximate continuous pseudo labels, the fine-grained inequality relationship can be inaccurate, while the coarse-grained inequality relationship can be relatively accurate. 
For the learning of the sample embedding space, the coarse-grained inequality relationship is more important, so we introduce a relaxation hyper-parameter $ \alpha $ so that our optimization goal only constrains the coarse-grained inequality relationship:
\begin{equation} \label{relaxation}
	\begin{aligned}
		\mathcal{L}_{\text{K}}
		= & \sum_{i=1}^{B}
		\sum_{j=1}^{B}
		\sum_{k=1}^{B}
		\mathbbm{1}(r_{ij} > r_{ik} + \alpha)
		\cdot
		\left[ 
		s(V_{i}, T_{k}) - s(V_{i}, T_{j}) 
		\right]_{+} \\
		+ & \sum_{i=1}^{B}
		\sum_{j=1}^{B}
		\sum_{k=1}^{B}
		\mathbbm{1}(r_{ji} > r_{ki} + \alpha)
		\cdot
		\left[ 
		s(V_{k}, T_{i}) - s(V_{j}, T_{i}) 
		\right]_{+}.
	\end{aligned}
\end{equation}
That is, if the difference between the correlation of two sample pairs is less than $ \alpha $, we no longer restrict their inequality relationship. 
When $ s(V_{i}, T_{j}) > s(V_{i}, T_{k}), \forall r_{ij} > r_{ik} + \alpha $ and $ s(V_{k}, T_{i}) > s(V_{j}, T_{i}), \forall r_{ji} > r_{ki} + \alpha $ are satisfied, $ \mathcal{L}_{\text{K}} $ is minimized.

For the selection of relaxation hyper-parameter $ \alpha $, we also designed a scheme. 
Since $ \alpha $ is introduced to avoid the negative impact of inaccurate continuous pseudo labels, we can count the standard deviation of the textual similarities between multiple captions of the same image as $ \alpha $.
The calculation method of textual similarity is the same as that of continuous pseudo labels.
For common image-text datasets, an image usually corresponds to multiple captions, for example, each image in Flickr30K and MS-COCO corresponds to $ 5 $ captions.
The standard deviation of textual similarities between multiple captions corresponding to the same image can reflect the error of the continuous pseudo labels of the same category.
According to our statistics, this standard deviation on the Flickr30K and MS-COCO datasets is around $ 0.2 $, so we set $ \alpha $ to $ 0.2 $ in the experiment. 
The experiments in SubSection \ref{parameter} prove that our hyper-parameter selection scheme can be flexibly applied to various data and retrieval models.

\begin{figure}[t]
	\centering
	\includegraphics[width=\linewidth]{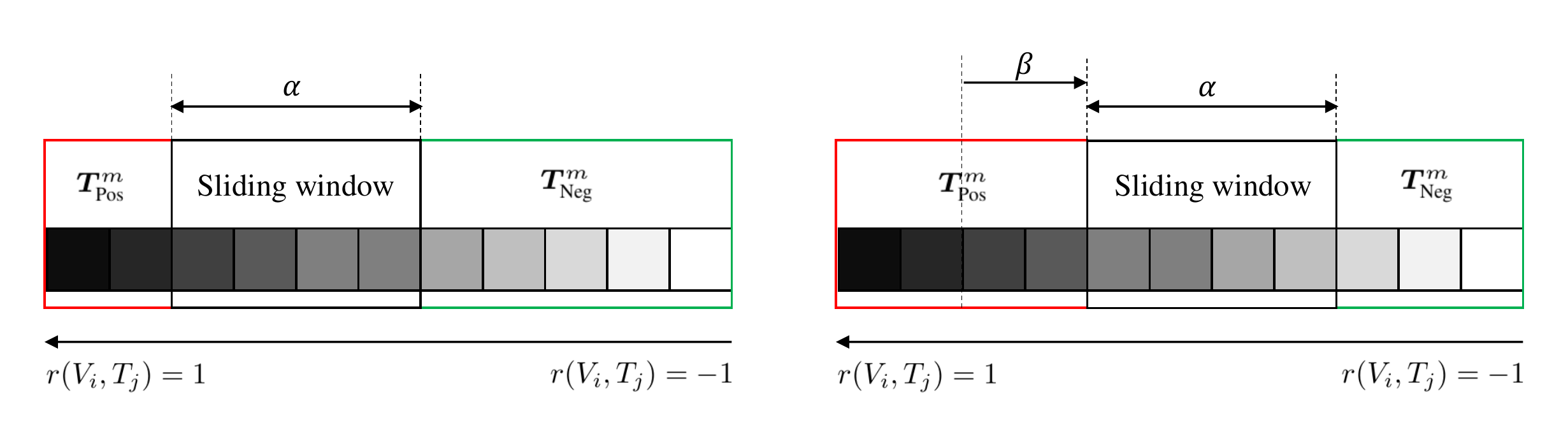}
	\caption{Sliding window sampling. The window size is the relaxation $ \alpha $ and the stride is $ \beta $.}
	\label{window}
\end{figure}
Calculating \textit{Kendall} $ \tau $ has high computational complexity. 
For a dataset with $ N $ samples, the computational complexity of calculating \textit{Kendall} $ \tau $ is $ O(N^{3}) $, which is difficult for deep learning training with a large batch. 
Therefore, we design a \textit{Sliding Window sampling and Hard Sample mining strategy} (\textit{SW-HS}) to reduce the complexity.

The sliding window sampling strategy is shown in the \figurename~\ref{window}, where the window size is the relaxation $ \alpha $ and the stride $ \beta $ is a hyper-parameter. 
Sliding window sampling is performed on the continuous pseudo labels between image-text samples. 
Each time a positive sample set $ \bm{T}^{m}_{\text{Pos}} $ (or $ \bm{V}^{m}_{\text{Pos}} $) and a negative sample set $ \bm{T}^{m}_{\text{Neg}} $ (or $ \bm{V}^{m}_{\text{Neg}} $) are sampled, the number of sampling $ M = (2 - \alpha)/\beta $ is determined by $ \alpha $ and $ \beta $.
In each sampling, the pseudo label of the positive sample set is greater than the negative sample set plus the relaxation $ \alpha $.
This is consistent with the constraints of Eq.~\eqref{relaxation}.
The similarity between samples in the positive sample set and anchor should be larger than samples in the negative sample set. 
We let the optimization objective only constrain the inequality relationship between the sampled positive and negative sample sets:
\begin{equation}
	\begin{aligned}
		\mathcal{L}_{\text{K-SW}}
		= & \sum_{m=1}^{M}
		\sum_{i=1}^{B} 
		\sum_{T_{j} \in \bm{T}^{m}_{\text{Pos}}}
		\sum_{T_{k} \in \bm{T}^{m}_{\text{Neg}}} 
		\left[ s(V_{i}, T_{k}) - s(V_{i}, T_{j}) \right]_{+} \\
		+ & \sum_{m=1}^{M}
		\sum_{i=1}^{B} 
		\sum_{V_{j} \in \bm{V}^{m}_{\text{Pos}}}
		\sum_{V_{k} \in \bm{V}^{m}_{\text{Neg}}}
		\left[ s(V_{k}, T_{i}) - s(V_{j}, T_{i}) \right]_{+}.
	\end{aligned}
\end{equation}
When $ s(V_{i}, T_{j}) > s(V_{i}, T_{k}), \forall T_{j} \in \bm{T}^{m}_{\text{Pos}}, T_{k} \in \bm{T}^{m}_{\text{Neg}} $ and $ s(V_{k}, T_{i}) > s(V_{j}, T_{i}), \forall V_{j} \in \bm{V}^{m}_{\text{Pos}}, V_{k} \in \bm{V}^{m}_{\text{Neg}} $ are satisfied, $ \mathcal{L}_{\text{K-SW}} $ can thus be minimized .

For image-text retrieval, the hard negative mining strategy \cite{faghri2018vse++}, where selected hard samples (instead of all samples) are utilized for the loss computation, has been shown to bring significant performance improvements.
Inspired by \cite{faghri2018vse++}, we develop a similar strategy of selecting hard samples for the \textit{Kendall ranking loss} computation. 
Instead of conducting the sum over the sampled positive and negative sample sets, we sample one pair $ (\check{T}_{i}^{m}, \hat{T}_{i}^{m}) $ (or $ (\check{V}_{i}^{m}, \hat{V}_{i}^{m}) $) from $ \bm{T}^{m}_{\text{Pos}} $ and $ \bm{T}^{m}_{\text{Neg}} $ (or $ \bm{V}^{m}_{\text{Pos}} $ and $ \bm{V}^{m}_{\text{Neg}} $) respectively, where $ \check{T}_{i}^{m} $ (or $ \check{V}_{i}^{m} $) is the furthest sample to the anchor $ V_{i} $ (or $ T_{i} $), $ \hat{T}_{i}^{m} $ (or $ \hat{V}_{i}^{m} $) is the closest sample to $ V_{i} $ (or $ T_{i} $). 
Finally, we average the results of $ M $ hard sample pairs. Thus, the \textit{Kendall rank loss} with \textit{Sliding Window sampling and Hard Sample mining strategy} (\textit{SW-HS}) can be rewritten as:
\begin{equation}
	\begin{aligned}
		\mathcal{L}_{\text{K-SW-HS}}
		= & \frac{1}{M} \sum_{m=1}^{M} \sum_{i=1}^{B} 
		\left[ s(V_{i}, \hat{T}_{i}^{m}) - s(V_{i}, \check{T}_{i}^{m}) \right]_{+}  \\
		+ & \frac{1}{M} \sum_{m=1}^{M} \sum_{i=1}^{B} 
		\left[ s(\hat{V}_{i}^{m}, T_{i}) - s(\check{V}_{i}^{m}, T_{i}) \right]_{+},
	\end{aligned}
\end{equation}
where
\begin{equation}
	\begin{aligned}
		\check{T}_{i}^{m}
		= & \arg \min _{T_{j} \in \bm{T}^{m}_{\text{Pos}}} 
		s(V_{i}, T_{j}), \\
		\hat{T}_{i}^{m}
		= & \arg \max _{T_{k} \in \bm{T}^{m}_{\text{Neg}}} 
		s(V_{i}, T_{k}), \\
		\check{V}_{i}^{m}
		= & \arg \min _{V_{j} \in \bm{V}^{m}_{\text{Pos}}} 
		s(V_{j}, T_{i}), \\
		\hat{V}_{i}^{m} 
		= & \arg \max _{V_{k} \in\bm{V}^{m}_{\text{Neg}}}  s(V_{k}, T_{i}).
	\end{aligned}
\end{equation}
With \textit{SW-HS}, the complexity of the \textit{Kendall rank loss} is reduced from $ O(N^{3}) $ to $ O(MN^{2}) $. Since the number of samples $ M $ is a constant much smaller than $ N $, it can be ignored. 
During training, we add $ \mathcal{L}_{\text{T-SN}} $ and $ \mathcal{L}_{\text{K-SW-HS}} $ as the final optimization objective of our framework with \textit{BCLS}.
The complexity of both loss functions is $ O(N^{2}) $.
Thus the complexity of our framework is $ O(N^{2}) $, which is the same magnitude as  the common triplet ranking loss.

\section{Experiments}\label{experiments}
\begin{table*}[t]
	\caption{Experimental results (\%) on Flickr30K and MS-COCO 1K. *: Ensemble results of two models.}
	\setlength\tabcolsep{4pt}
	\begin{center}
		\begin{tabular}{cccccccccccccccc}
			\toprule[1pt]
			Data Split
			& \multicolumn{7}{c}{Flickr30K 1K Test} & & \multicolumn{7}{c}{MS-COCO 5-fold 1K Test} \\
			\cline{2-8}\cline{10-16}
			\specialrule{0em}{2pt}{0pt}
			Eval Task 
			& \multicolumn{3}{c}{Image-to-Text} & \multicolumn{3}{c}{Text-to-Image} & \multirow{2}*{RSUM} & & \multicolumn{3}{c}{Image-to-Text} & \multicolumn{3}{c}{Text-to-Image} & \multirow{2}*{RSUM} \\
			\cline{2-7}\cline{10-15}
			\specialrule{0em}{2pt}{0pt}
			Method 
			& R@1 & R@5 & R@10 & R@1 & R@5 & R@10 & & & R@1 & R@5 & R@10 & R@1 & R@5 & R@10 \\
			\hline

			\specialrule{0em}{2pt}{0pt}
			SAF$ _{\text{(\textit{AAAI}'21)}} $ \cite{diao2021similarity} & 73.7 & 93.3 & 96.3 & 56.1 & 81.5 & 88.0 & 488.9 & & 76.1 & 95.4 & 98.3 & 61.8 & 89.4 & 95.3 & 516.3 \\
			\rowcolor{black!10}
			\textbf{SAF + \textit{BCLS}} & \textbf{77.9} & \textbf{95.0} & \textbf{97.4} & \textbf{57.7} & \textbf{83.9} & \textbf{89.5} & \textbf{501.3} & & \textbf{77.7} & \textbf{95.8} & \textbf{98.5} & \textbf{63.0} & \textbf{89.9} & \textbf{95.5} & \textbf{520.4} \\
			
			\hline
			\specialrule{0em}{2pt}{0pt}
			SGR$ _{\text{(\textit{AAAI}'21)}} $ \cite{diao2021similarity} &  75.2 & 93.3 & 96.6 & 56.2 & 81.0 & 86.5 & 488.8 & & 78.0 & 95.8 & 98.2 & 61.4 & 89.3 & \textbf{95.4} & 518.1 \\
			\rowcolor{black!10}
			\textbf{SGR + \textit{BCLS}} & \textbf{78.2} & \textbf{94.7} & \textbf{97.7} & \textbf{58.1} & \textbf{83.7} & \textbf{89.1} & \textbf{501.6} & & \textbf{78.1} & \textbf{96.2} & \textbf{98.7} & \textbf{63.0} & \textbf{89.9} & 95.0 & \textbf{520.8} \\
			
			\hline
			\specialrule{0em}{2pt}{0pt}
			SGRAF*$ _{\text{(\textit{AAAI}'21)}} $ \cite{diao2021similarity} & 77.8 & 94.1 & 97.4 & 58.5 & 83.0 & 88.8 & 499.6 & & 79.6 & 96.2 & 98.5 & 63.2 & 90.7 & \textbf{96.1} & 524.3 \\
			\rowcolor{black!10}
			\textbf{SGRAF* + \textit{BCLS}} & \color{red}\textbf{81.0} & \color{red}\textbf{95.7} & \color{red}\textbf{98.0} & \color{red}\textbf{61.0} & \color{red}\textbf{85.4} & \color{red}\textbf{90.4} & \color{red}\textbf{511.5} & & \color{red}\textbf{80.3} & \color{red}\textbf{96.6} & \color{red}\textbf{99.0} & \color{red}\textbf{64.5} & \color{red}\textbf{90.9} & 95.8 & \color{red}\textbf{527.1} \\
			
			\bottomrule[1pt]
		\end{tabular}
	\end{center}
	\label{f30kcoco1k}
\end{table*}
\subsection{Dataset and Experiment Settings} \label{dataset}
We evaluate our method on two standard benchmarks: Flickr30K \cite{young2014image} and MS-COCO \cite{lin2014microsoft}. 
Flickr30K dataset contains 31,000 images, each image is annotated with 5 sentences. 
Following the data split of \cite{lee2018stacked}, we use 1,000 images for validation, 1,000 images for testing, and the remaining for training. 
MS-COCO dataset contains 123,287 images, and each image comes with 5 sentences. 
We mirror the data split setting of \cite{lee2018stacked}. 
More specifically, we use 113,287 images for training, 5,000 images for validation, and 5,000 images for testing. 
We report results on both 1,000 test images (averaged over 5 folds) and full 5,000 test images of MS-COCO.

Our all experiments are done on an NVIDIA GeForce RTX 3090 GPU using PyTorch 1.7.1. 
Models are trained using Adam for 20 epochs, with a batch size of 128 for both datasets. 
The initial learning rate of these two models is set as 0.0005 for the first 10 epochs and then decays by a factor of 10 for the last 10 epochs. 
Hyper-parameters are set as $ \alpha = 0.2 $, $ \beta = 0.1 $ and $ \gamma = 50 $, for both Flickr30K and MS-COCO.

\subsection{Evaluation Metric}
For the evaluation of instance-based retrieval, following the \cite{faghri2018vse++}, we use the Recall@K (R@K), with $ K = \{1,5,10\} $ as the evaluation metric for the task. 
R@K indicates the percentage of queries for which the model returns the correct item in its top $ K $ results. 
We follow \cite{chen2021learning} to use RSUM, which is defined as the sum of recall metrics at $ K = \{1,5,10\} $ of both text-to-image and image-to-text retrievals, as an average metric to gauge retrieval model's overall performances.

For the evaluation of semantic-based retrieval, existing work proposes to use nDCG \cite{wray2021semantic}, Coherent Score \cite{zhou2020ladder, wang2021adaptive} and Normalized Cumulative Semantic Score \cite{biten2021image} to evaluate the performance of semantic-based retrieval methods.
But these evaluation metrics are all based on inaccurate pseudo labels. 
The evaluation results will result in poor overfitting since the pseudo labels of the same distribution are used for training and testing.
The evaluation metrics can only reflect the fit of the retrieval model to inaccurate pseudo labels, which cannot objectively reflect the retrieval performance.
To address the above issues, we choose to perform a performance evaluation of semantic-based retrieval methods on the Extended COCO Validation (ECCV) Caption dataset \cite{chun2022eccv}. 
The ECCV Caption dataset extends the test set of the MS-COCO dataset with the machine and human annotations, providing semantic associations for more image-text pairs.
We use the MS-COCO training set to train the model and use the ECCV Caption dataset for testing.
Besides R@K, following Chun \textit{et al.} \cite{chun2022eccv}, we use mAP@R and R-Precision (R-P) to test the performance of existing semantic-based retrieval methods. 
Since the ECCV Caption dataset is annotated by machines and humans, the evaluation metrics used for testing are not based on approximate pseudo labels, so the performance of semantic-based retrieval methods can be compared more objectively and fairly.

\subsection{Comparisons with State-of-the-art Methods}
\begin{table}[t] 
	\caption{Experimental results (\%) on MS-COCO 5K. *: Ensemble results of two models}
	\setlength\tabcolsep{3pt}
	\begin{center}
		\begin{tabular}{ccccccccccccc}
			\toprule[1pt]
			Eval Task 
			& \multicolumn{3}{c}{Image-to-Text} & \multicolumn{3}{c}{Text-to-Image} & \multirow{2}*{RSUM}\\
			\cline{2-7}
			\specialrule{0em}{2pt}{0pt}
			Method & R@1 & R@5 & R@10 & R@1 & R@5 & R@10 \\
			\hline
			
			\specialrule{0em}{2pt}{0pt}
			SCAN* \cite{lee2018stacked} & 50.4 & 82.2 & 90.0 & 38.6 & 69.3 & 80.4 & 410.9 \\
			VSRN* \cite{li2019visual} & 53.0 & 81.1 & 89.4 & 40.5 & 70.6 & 81.1 & 415.7 \\
			IMRAM* \cite{chen2020imram} & 53.7 & 83.2 & 91.0 & 39.7 & 69.1 & 79.8 & 416.5 \\
			PFAN++* \cite{wang2020pfan++} & 51.2 & 84.3 & 89.2 & 41.4 & 70.9 & 79.0 & 416.0 \\
			VSE$\infty$ \cite{chen2021learning} & 56.6 & 83.6 & 91.4 & 39.3 & 69.9 & 81.1 & 421.9 \\
			UARDA* \cite{zhang2022unified} & 56.2 & 83.8 & 91.3 & 40.6 & 69.5 & 80.9 & 422.3 \\
			
			\hline
			\specialrule{0em}{2pt}{0pt}
			SAF \cite{diao2021similarity} & 53.3 & - & 90.1 & 39.8 & - & 80.2 & - \\
			\rowcolor{black!10}
			\textbf{SAF + \textit{BCLS}} & \textbf{54.8} & \textbf{83.6} & \textbf{91.6} & \textbf{41.3} & \textbf{70.5} & \textbf{80.9} & \textbf{422.6} \\
			
			\hline
			\specialrule{0em}{2pt}{0pt}
			SGR \cite{diao2021similarity} & \textbf{56.9} & - & 90.5 & 40.2 & - & 79.8 & - \\
			\rowcolor{black!10}
			\textbf{SGR + \textit{BCLS}} &  56.2 & \textbf{84.3} & \textbf{91.4} & \textbf{41.2} & \textbf{70.5} & \textbf{80.9} & \textbf{424.5} \\
			
			\hline
			\specialrule{0em}{2pt}{0pt}
			SGRAF* \cite{diao2021similarity} & 57.8 & - & 91.6 & 41.9 & - & 81.3 & - \\
			\rowcolor{black!10}
			\textbf{SGRAF* + \textit{BCLS}} & \color{red}\textbf{59.5} & \color{red}\textbf{85.5} & \color{red}\textbf{92.4} & \color{red}\textbf{43.1} & \color{red}\textbf{72.1} & \color{red}\textbf{82.3} & \color{red}\textbf{434.9} \\
			
			\bottomrule[1pt]
		\end{tabular}
	\end{center}
\label{coco5k}
\end{table}
Since the framework with \textit{BCLS} proposed in this paper is independent of image and text feature representation and similarity calculation, and can be plug-and-play applied to existing instance-based retrieval models.
We apply the proposed framework to the state-of-the-art model SGRAF \cite{diao2021similarity}, denoted as \textbf{\textit{+ BCLS}}.
SGRAF is a similarity graph reasoning and attention filtration network for image-text matching, which is currently the best performing open source image-text retrieval method without pre-training. 
SGRAF consists of two models, SGR and SAF, and we report the performance of our framework on both models as well as the performance of the ensemble model.
We compare our method with recent state-of-the-art methods on Flickr30K and MS-COCO datasets. 
For a fair comparison, the feature extraction backbone of all methods is the same, \textit{i.e.}, that for image is Faster R-CNN \cite{anderson2018bottom}, and that for text is Bi-directional GRU (Bi-GRU) \cite{schuster1997bidirectional}.

\tablename~\ref{f30kcoco1k} compares our method with state-of-the-art image-text retrieval methods on Flickr30K and MS-COCO 1K test set. 
Compared with the baselines, the retrieval models with our framework can achieve better performance in almost all evaluation metrics.
On the Flickr30K dataset, \textbf{SGRAF* + \textit{BCLS}} improves RSUM by 11.9\% compared to SGRAF*. 
On MS-COCO 1K test set, applying the proposed framework to SGRAF* can improve RSUM by 2.8\%.
As shown in \tablename~\ref{coco5k}, our method boosts the performance of almost all evaluation metrics on both baselines on the MS-COCO 5K test set.
Our method yields a 1.7\% increase for image-to-text retrieval and 1.1\% improvement for text-to-image retrieval in terms of R@1 when compared with SGRAF*.
Our framework can guide the model to learn continuous semantic relations, thus making the learned embedding space more reasonable. 
Experiments show that our method can further improve the performance of existing state-of-the-art models.
Our framework with \textit{BCLS} can achieve significant performance improvements on multiple models and datasets using the same set of hyper-parameters, indicating that our framework can be flexibly applied to various data and retrieval models without complex parameter settings.

\subsection{Comparison with Semantic-based Retrieval Methods}
\begin{table*}[t]
	\caption{Experimental results (\%) on Flickr30K and MS-COCO 5K.}
	\setlength\tabcolsep{4pt}
	\begin{center}
		\begin{tabular}{cccccccccccccccc}
			\toprule[1pt]
			Data Split
			& \multicolumn{7}{c}{Flickr30K 1K Test} & & \multicolumn{7}{c}{MS-COCO 5K Test} \\
			\cline{2-8}\cline{10-16}
			\specialrule{0em}{2pt}{0pt}
			Eval Task 
			& \multicolumn{3}{c}{Image-to-Text} & \multicolumn{3}{c}{Text-to-Image} & \multirow{2}*{RSUM} & & \multicolumn{3}{c}{Image-to-Text} & \multicolumn{3}{c}{Text-to-Image} & \multirow{2}*{RSUM} \\
			\cline{2-7}\cline{10-15}
			\specialrule{0em}{2pt}{0pt}
			Method 
			& R@1 & R@5 & R@10 & R@1 & R@5 & R@10 & & & R@1 & R@5 & R@10 & R@1 & R@5 & R@10 \\
			\hline
			
			\specialrule{0em}{2pt}{0pt}
			VSE++ (ResNet) & 48.9 & 77.8 & 86.5 & 36.0 & 65.8 & 75.4 & 390.4 & & 36.3 & 65.9 & 78.6 & 25.5 & 53.5 & 66.5 & 326.3 \\
			CVSE++ (ResNet) \cite{wang2021adaptive} & 48.9 & 77.8 & 86.7 & 36.1 & 66.3 & 75.5 & 391.3 & & \textbf{39.0} & 67.1 & 79.1 & 25.1 & 52.8 & 65.5 & 328.6 \\
			CVSE++ (ResNet, Auto) \cite{wang2021adaptive} & 49.3 & 77.2 & 86.3 & 36.0 & 65.4 & 76.2 & 390.4 & & 38.8 & 65.8 & 79.0 & 24.4 & 52.5 & 65.1 & 325.6\\
			\rowcolor{black!10}
			\textbf{VSE++ (ResNet) + \textit{BCLS}} & \textbf{54.4} & \textbf{82.8} & \textbf{89.5} & \textbf{39.9} & \textbf{69.9} & \textbf{79.1} & \textbf{415.6} & & 37.4 & \textbf{69.0} & \textbf{80.6} & \textbf{25.9} & \textbf{54.5} & \textbf{67.8} & \textbf{335.2} \\
			\hline
			
			\specialrule{0em}{2pt}{0pt}
			SGR \cite{diao2021similarity} &  75.2 & 93.3 & 96.6 & 56.2 & 81.0 & 86.5 & 488.8 & & \textbf{56.9} & - & 90.5 & 40.2 & - & 79.8 & - \\
			SGR + SAM \cite{biten2021image} & 75.9 & 92.4 & 96.6 & 57.6 & 83.1 & \textbf{89.7} & 495.3 & & 55.7 & 83.2 & 91.2 & 40.5 & 69.7 & 80.5 & 420.8 \\
			\rowcolor{black!10}
			\textbf{SGR + \textit{BCLS}} & \textbf{78.2} & \textbf{94.7} & \textbf{97.7} & \textbf{58.1} & \textbf{83.7} & 89.1 & \textbf{501.6} & & 56.2 & \textbf{84.3} & \textbf{91.4} & \textbf{41.2} & \textbf{70.5} & \textbf{80.9} & \textbf{424.5} \\
			
			\bottomrule[1pt]
		\end{tabular}
	\end{center}
	\label{f30kcoco5k}
\end{table*}
In this section, we conduct a fair comparison of existing semantic-based retrieval models:
\begin{itemize}
	\item \textbf{CVSE++:} CVSE++ adopt a ladder loss \cite{zhou2020ladder, wang2021adaptive} to learn a coherent embedding space. In ladder loss, relevance degrees are divided into several levels.
	\item \textbf{SAM:} SAM loss \cite{biten2021image} is a variant of triplet loss, candidates are pushed away from the query by semantic adaptive margins in the embedding space.
\end{itemize}

For a fair comparison, we used the same experimental setup as CVSE++ and SAM. Compared to CVSE++, we also apply our framework to the VSE++ model. A ResNet-152 pre-trained on ImageNet is used for image representation. 
Following \cite{zhou2020ladder, wang2021adaptive}, we also adopt random cropping in data augmentation, where all images are first resized to $ 256 \times 256 $ and randomly cropped 10 times at $ 224 \times 224 $ resolution. A GRU is used for text representation. 
Compared to SAM, we also apply our framework to the SGR model.

\tablename~\ref{f30kcoco5k} compares our method with existing semantic-based retrieval models on Flickr30K and MS-COCO 5K test set. 
Compared with the baselines, the retrieval models with our framework can achieve better performance in almost all evaluation metrics.
On the Flickr30K dataset, using our framework improves RSUM by 25.2\% compared to the original VSE++ model and 24.3\% compared to the CVSE++ model.
On the MS-COCO 5K test set, compared to VSE++ and CVSE++, our method is improved by 8.9\% and 6.6\% on RSUM, respectively.
Compared to SAM, our method can also boost RSUM by 6.3\% and 3.7\% on the Flickr30K and MS-COCO 5K test set datasets, respectively.

\begin{table}[t]
	\caption{Kendall $ \tau $ on Flickr30K and MS-COCO 5K.}
	\begin{center}
		\begin{tabular}{ccccccccccccc}
			\toprule[1pt]
			\multirow{2}*{Method} & \multicolumn{2}{c}{Flickr30K} & & \multicolumn{2}{c}{MS-COCO 5K} \\
			\cline{2-3}\cline{5-6}
			\specialrule{0em}{2pt}{0pt}
			~ & I$ \rightarrow $T & T$ \rightarrow $I &  & I$ \rightarrow $T & T$ \rightarrow $I \\
			\hline
			
			\specialrule{0em}{2pt}{0pt}
			VSE++ & 0.238 & 0.237 & & 0.129 & 0.138 \\
			CVSE++ & 0.238 & 0.237 & & 0.152 & 0.163 \\
			VSE++ + SAM & 0.256 & 0.258 & & 0.199 & 0.208 \\
			\rowcolor{black!10}
			\textbf{VSE++ + \textit{BCLS}} &  \textbf{0.291} & \textbf{0.287} & & \textbf{0.417} & \textbf{0.406} \\
			
			\bottomrule[1pt]
		\end{tabular}
	\end{center}
	\label{kendallf30kcoco}
\end{table}
\tablename~\ref{kendallf30kcoco} compares the \textit{Kendall} $ \tau $ of our method with existing semantic-based retrieval methods on the Flickr30K dataset and MS-COCO 5K test set. 
I$ \rightarrow $T means retrieval from image to text, T$ \rightarrow $I means retrieval from text to image.
\textit{Kendall} $ \tau $ is calculated based on continuous pseudo labels, which can reflect how well the model fits continuous pseudo labels.
It can be seen that the \textit{Kendall} $ \tau $ of our method on the test set is much higher than other methods since our \textit{Kendall ranking loss} is designed based on the definition of \textit{Kendall} $ \tau $, which can fully learn coherent semantic relations from continuous pseudo labels.
But using \textit{Kendall} $ \tau $ as an evaluation metric has drawbacks. \textit{Kendall} $ \tau $ is an evaluation metric based on pseudo labels, which can only reflect how well the model fits continuous pseudo labels but cannot objectively reflect retrieval performance.

\begin{table}[t]
	\caption{Experimental results (\%) on ECCV Caption.}
	\setlength\tabcolsep{4pt}
	\begin{center}
		\begin{tabular}{ccccccccccccc}
			\toprule[1pt]
			\multirow{2}*{Method} & \multicolumn{3}{c}{Image-to-Text} & \multicolumn{3}{c}{Text-to-Image} \\
			\cline{2-7}
			\specialrule{0em}{2pt}{0pt}
			~ & mAP@R & R-P & R@1 & mAP@R & R-P & R@1 \\
			\hline
			
			\specialrule{0em}{2pt}{0pt}
			VSE++ &  20.8 & 32.8 & 55.8 & 38.3 & 48.1 & 73.3 \\
			CVSE++ & 21.2 & 33.0 & 57.8 & 38.5 & 48.2 & 75.0 \\
			VSE++ + SAM & 21.4 & 33.5 & 55.4 & 38.4 & 48.3 & 76.1 \\
			\rowcolor{black!10}
			\textbf{VSE++ + \textit{BCLS}} & \textbf{21.8} & \textbf{33.8} & \textbf{59.3} & \textbf{39.2} & \textbf{49.0} & \textbf{76.3} \\
			
			\bottomrule[1pt]
		\end{tabular}
	\end{center}
	\label{eccv}
\end{table}
For the evaluation problem of semantic-based retrieval methods, we conducted an objective and fair comparison of existing semantic-based retrieval methods with the help of the ECCV Caption dataset. 
As shown in \tablename~\ref{eccv}, our method boosts the performance of all evaluation metrics on both baselines on the ECCV Caption dataset.
Our method yields a 1.5\% increase for image-to-text retrieval and 1.3\% improvement for text-to-image retrieval in terms of R@1 when compared with CVSE++.
Since \textit{Kendall ranking loss} helps the model learn a more coherent embedding space, where candidates with higher relevance degrees are mapped closer to the query than those with lower relevance degrees.
The comparison of semantic-based retrieval methods in this subsection is based on the objectively annotated ECCV Caption dataset, verifying that our method can better learn continuous semantic relations.

\subsection{Ablation Study}
\begin{table*}[t]
	\caption{Ablation studies on Flickr30K and MS-COCO 5K.}
	\setlength\tabcolsep{4pt}
	\begin{center}
		\begin{tabular}{cccccccccccccccc}
			\toprule[1pt]
			Data Split
			& \multicolumn{7}{c}{Flickr30K 1K Test} & & \multicolumn{7}{c}{MS-COCO 5K Test} \\
			\cline{2-8}\cline{10-16}
			\specialrule{0em}{2pt}{0pt}
			Eval Task 
			& \multicolumn{3}{c}{Image-to-Text} & \multicolumn{3}{c}{Text-to-Image} & \multirow{2}*{RSUM} & & \multicolumn{3}{c}{Image-to-Text} & \multicolumn{3}{c}{Text-to-Image} & \multirow{2}*{RSUM} \\
			\cline{2-7}\cline{10-15}
			\specialrule{0em}{2pt}{0pt}
			Method 
			& R@1 & R@5 & R@10 & R@1 & R@5 & R@10 & & & R@1 & R@5 & R@10 & R@1 & R@5 & R@10 \\
			\hline
			
			\specialrule{0em}{2pt}{0pt}
			$ \mathcal{L}_{\text{T-HN}} $ & 48.9 & 77.8 & 86.5 & 36.0 & 65.8 & 75.4 & 390.4 & & 36.3 & 65.9 & 78.6 & 25.5 & 53.5 & 66.5 & 326.3 \\
			\textit{BCLS} (w/o $ \mathcal{L}_{\text{T-SN}} $) & 42.6 & 73.2 & 83.8 & 30.3 & 61.1 & 72.6 & 363.5 & & 22.7 & 48.1 & 61.5 & 14.9 & 37.6 & 50.7 & 235.5 \\
			\textit{BCLS} (w/o $ \mathcal{L}_{\text{K-SW-HS}} $) & 52.0 & 79.2 & 87.9 & 36.9 & 66.9 & 77.3 & 400.1 & & 37.0 & 66.9 & 79.0 & 25.5 & 53.9 & 67.0 & 329.4 \\
			\textbf{\textit{BCLS}} & \textbf{54.4} & \textbf{82.8} & \textbf{89.5} & \textbf{39.9} & \textbf{69.9} & \textbf{79.1} & \textbf{415.6} & & \textbf{37.4}
			& \textbf{69.0} & \textbf{80.6} & \textbf{25.9} & \textbf{54.5} & \textbf{67.8} & \textbf{335.2} \\
			\bottomrule[1pt]
		\end{tabular}
	\end{center}
\label{ablation}
\end{table*}
\begin{table}[t]
	\caption{Ablation studies on ECCV Caption.}
	\setlength\tabcolsep{4pt}
	\begin{center}
		\begin{tabular}{ccccccccccccc}
			\toprule[1pt]
			\multirow{2}*{Method} & \multicolumn{3}{c}{Image-to-Text} & \multicolumn{3}{c}{Text-to-Image} \\
			\cline{2-7}
			\specialrule{0em}{2pt}{0pt}
			~ & mAP@R & R-P & R@1 & mAP@R & R-P & R@1 \\
			\hline
			
			\specialrule{0em}{2pt}{0pt}
			$ \mathcal{L}_{\text{T-HN}} $ &  20.8 & 32.8 & 55.8 & 38.3 & 48.1 & 73.3 \\
			\textit{BCLS} (w/o $ \mathcal{L}_{\text{T-SN}} $) & 13.3 & 23.9 & 38.6 & 25.7 & 36.6 & 53.8 \\
			\textit{BCLS} (w/o $ \mathcal{L}_{\text{K-SW-HS}} $) & 21.2 & 33.1 & 57.8 & 38.3 & 47.9 & 75.8 \\
			\textit{\textbf{BCLS}} & \textbf{21.8} & \textbf{33.8} & \textbf{59.3} & \textbf{39.2} & \textbf{49.0} & \textbf{76.3} \\
			
			\bottomrule[1pt]
		\end{tabular}
	\end{center}
\label{ablationeccv}
\end{table}
In order to verify the effect of each part of our framework with \textit{BCLS}, we conduct a comprehensive ablation study on the VSE++ model.
\tablename~\ref{ablation} shows the ablation experimental results on Flickr30K and MS-COCO 5K test set.
\textit{BCLS} (w/o $ \mathcal{L}_{\text{T-SN}} $) means using $ \mathcal{L}_{\text{K-SW-HS}} $ alone without $ \mathcal{L}_{\text{T-SN}} $, which cannot achieve good performance. since $ \mathcal{L}_{\text{K-SW-HS}} $ does not bring good discrimination to binary labeled positive and negative samples. 
\textit{BCLS} (w/o $ \mathcal{L}_{\text{K-SW-HS}} $) means using $ \mathcal{L}_{\text{T-SN}} $ alone without $ \mathcal{L}_{\text{K-SW-HS}} $.
It can be seen that our improved $ \mathcal{L}_{\text{T-SN}} $ can achieve better performance than the most widely used $ \mathcal{L}_{\text{T-HN}} $. 
Jointly optimizing $ \mathcal{L}_{\text{T-SN}} $ and $ \mathcal{L}_{\text{K-SW-HS}} $ (\textit{\textbf{BCLS}}) can achieve the best performance. 
This shows that the learning of continuous semantic relations by $ \mathcal{L}_{\text{K-SW-HS}} $ also has a positive impact on instance-based retrieval performance.
As shown in \tablename~\ref{ablationeccv}, jointly optimizing $ \mathcal{L}_{\text{T-SN}} $ and $ \mathcal{L}_{\text{K-SW-HS}} $ (\textit{\textbf{BCLS}}) can bring significant performance improvement on the ECCV Caption dataset. This shows that $ \mathcal{L}_{\text{K-SW-HS}} $ can guide the model to learn a more coherent embedding space.

\begin{figure}[t]
	\centering
	\subfigure[Loss]{
		\label{train_loss}
		\includegraphics[width=0.46\linewidth]{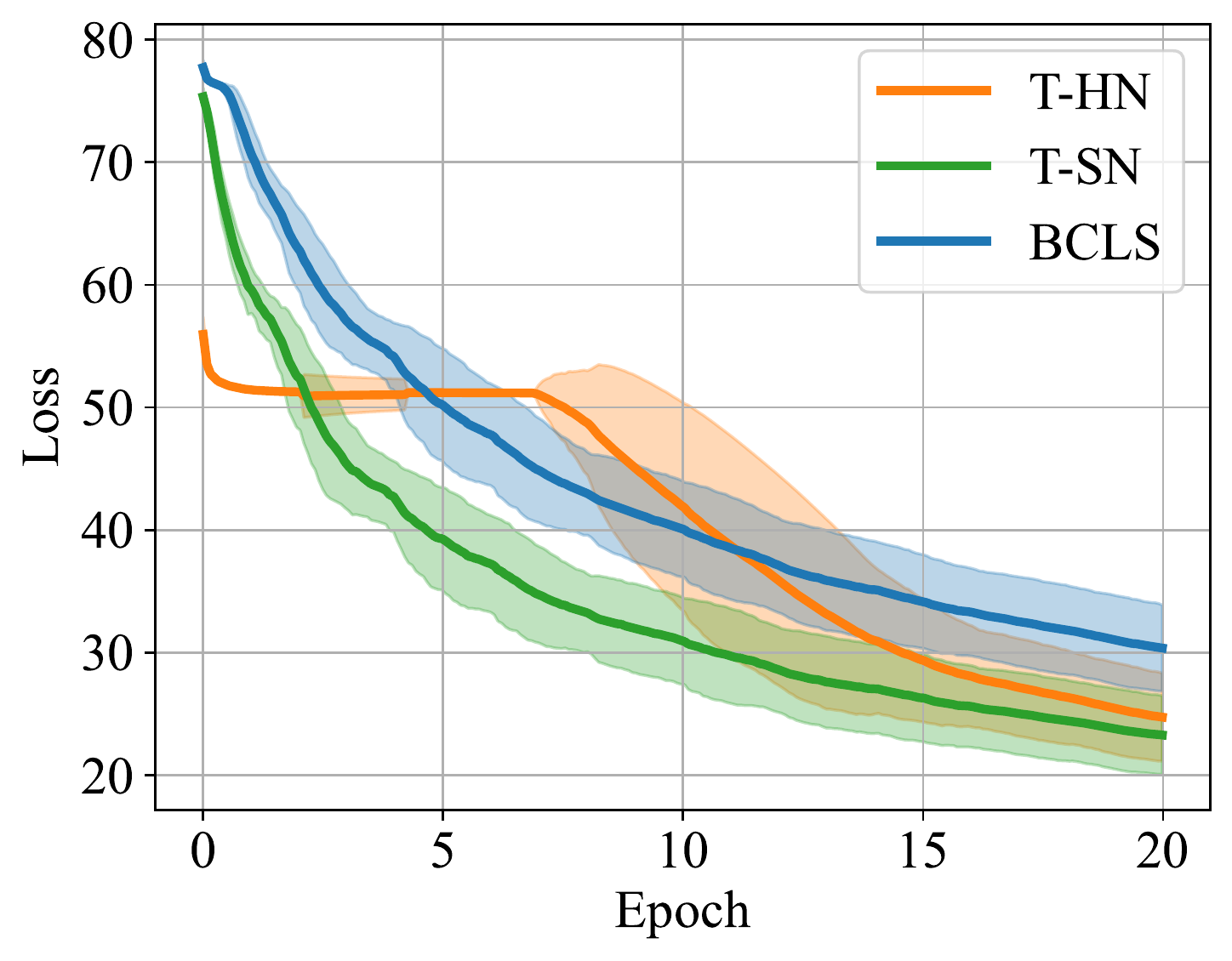}
	}
	\subfigure[RSUM]{
		\label{train_rsum}
		\includegraphics[width=0.46\linewidth]{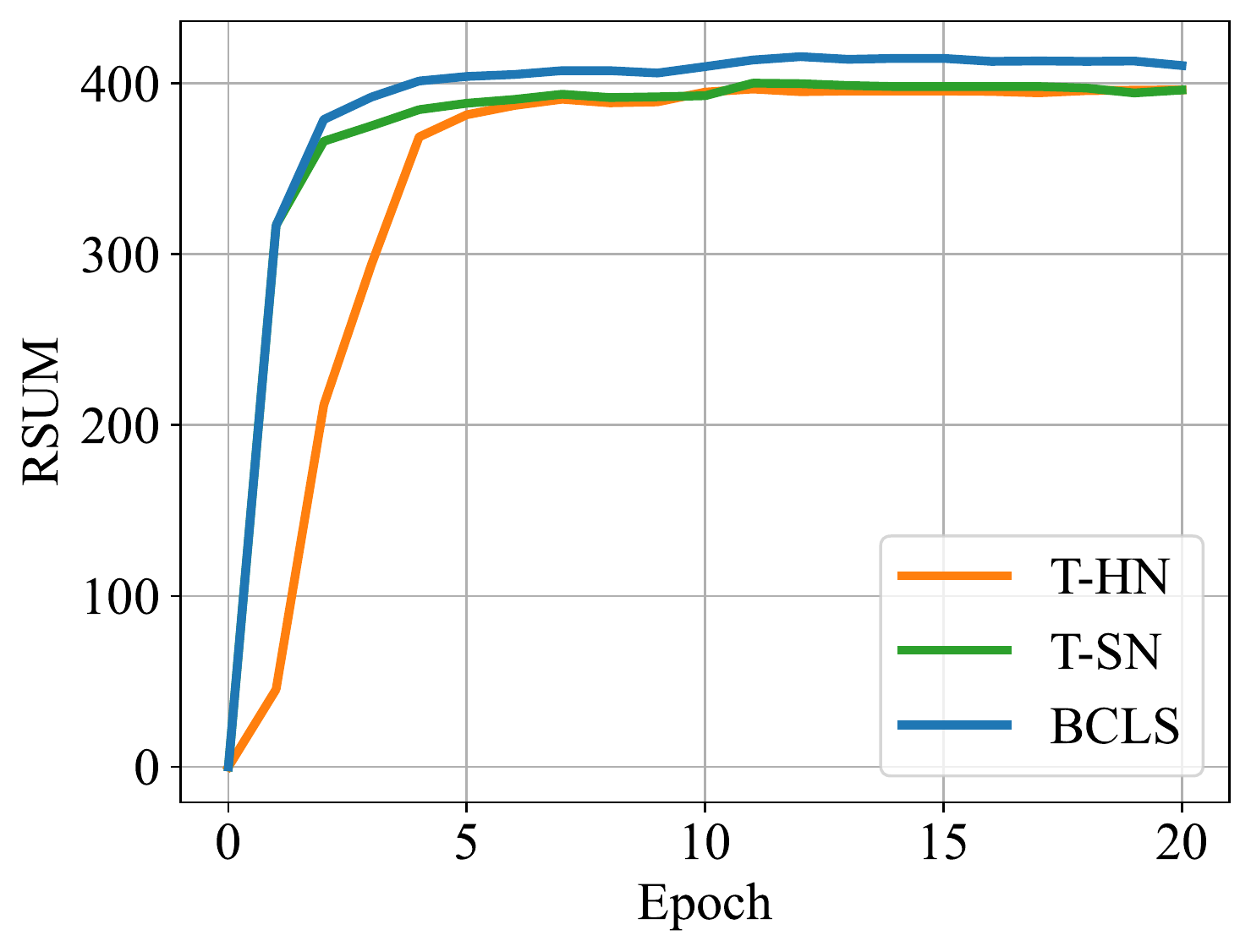}
	}
	\caption{Plotting training epoch against loss and RSUM on Flickr30K validation set.}
\end{figure}
\figurename~\ref{train_loss} and \figurename~\ref{train_rsum} compare the performance of $ \mathcal{L}_{\text{T-HN}} $, $ \mathcal{L}_{\text{T-SN}} $ alone and using our framework with \textit{BCLS} during training.
We plot training epoch against loss and RSUM on Flickr30K validation set.
It can be seen from \figurename~\ref{train_loss} that our improved $ \mathcal{L}_{\text{T-SN}} $ has better convergence than $ \mathcal{L}_{\text{T-HN}} $. 
$ \mathcal{L}_{\text{T-SN}} $ also reduces the optimization difficulty of our complete framework.
\figurename~\ref{train_rsum} shows that the retrieval model can achieve superior retrieval performance faster whether using our proposed $ \mathcal{L}_{\text{T-SN}} $ alone or using the framework with \textit{BCLS}.

\subsection{Parameter Analysis}
\label{parameter}
\begin{figure}[t]
	\centering
	\subfigure[RSUM, relaxation $ \alpha $]{
		\label{gap_rsum}
		\includegraphics[width=0.46\linewidth]{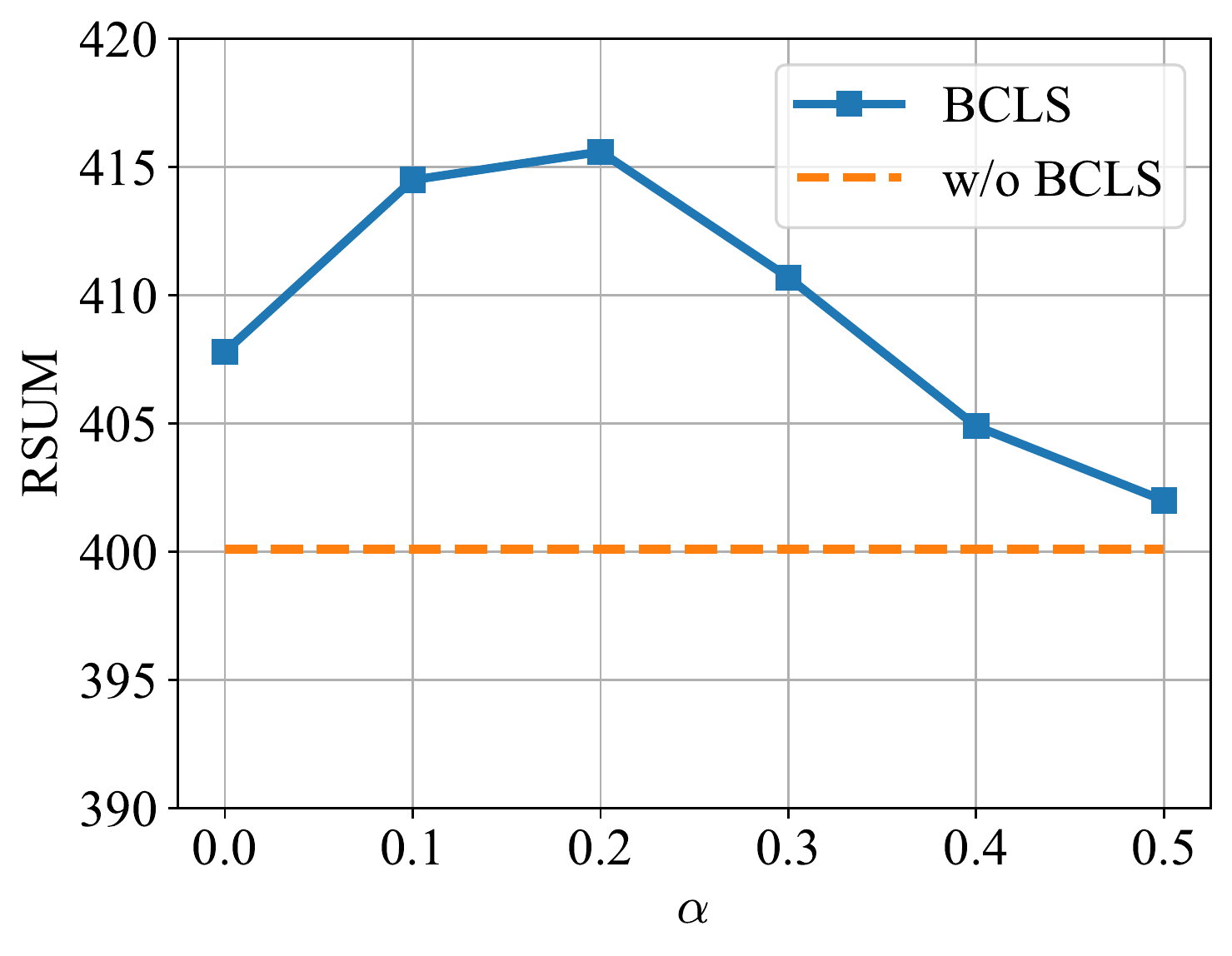}
	}
	\subfigure[Kendall $ \tau $, relaxation $ \alpha $]{
		\label{gap_kendall}
		\includegraphics[width=0.46\linewidth]{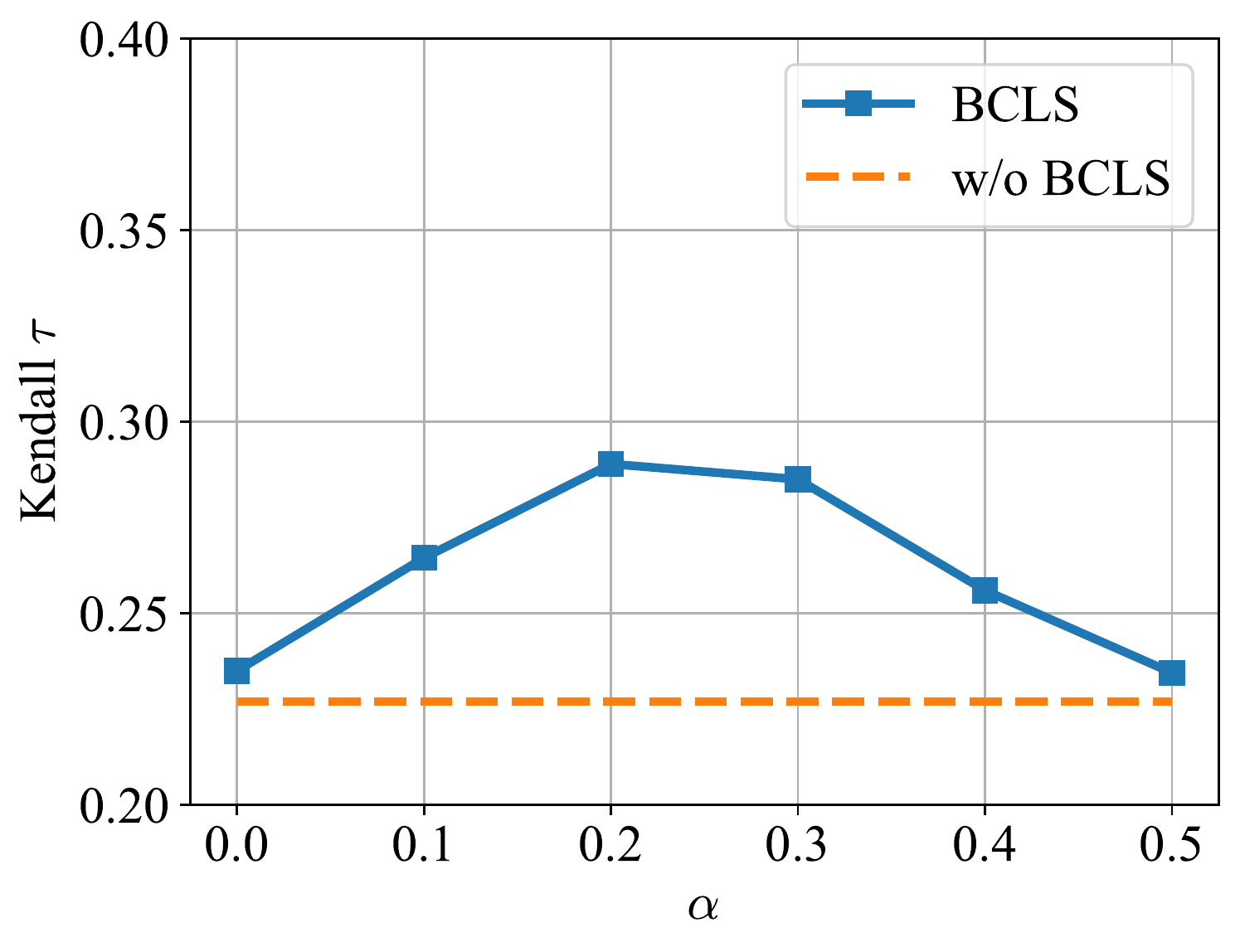}
	}
	\subfigure[RSUM, stride $ \beta $]{
		\label{stride_rsum}
		\includegraphics[width=0.46\linewidth]{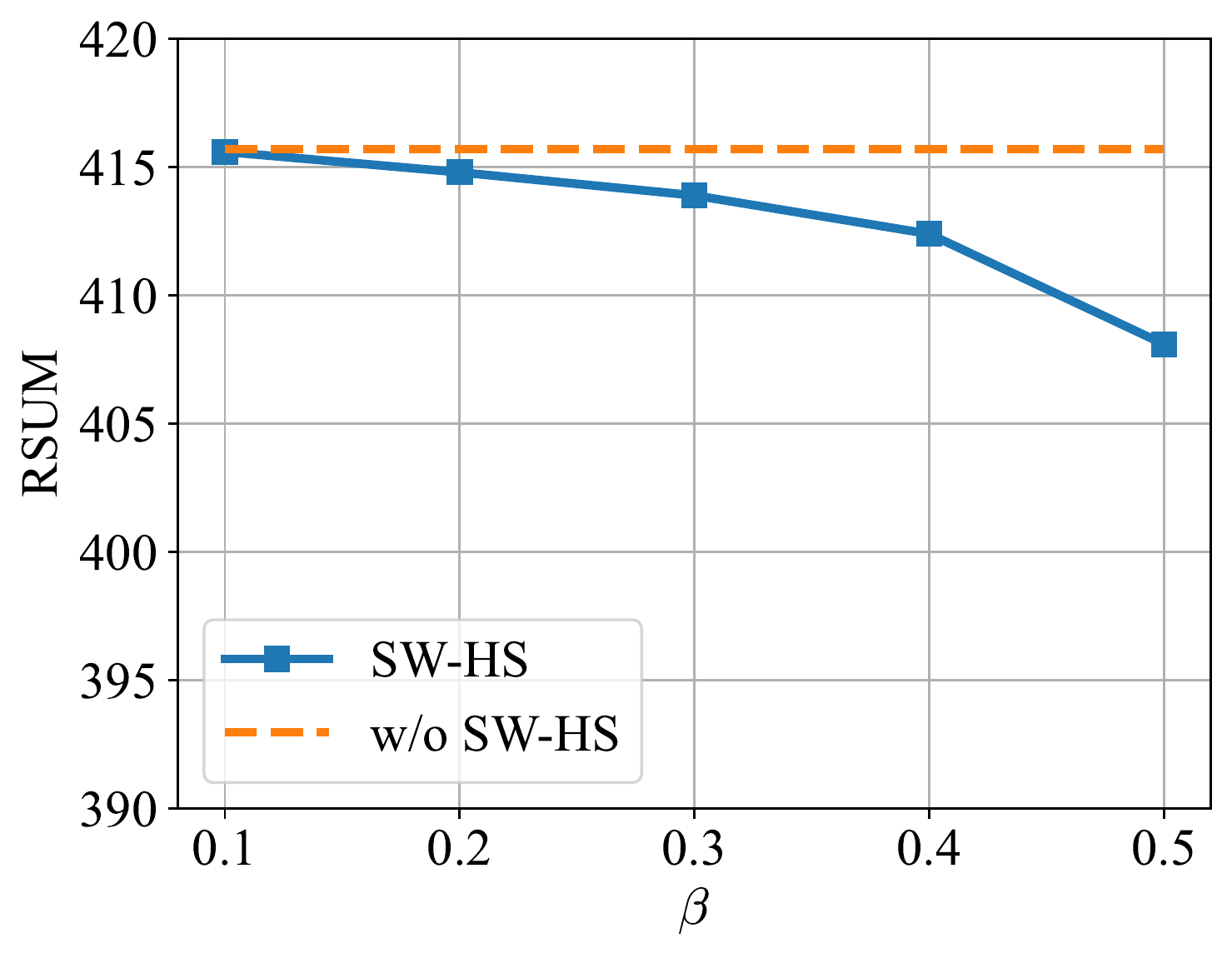}
	}
	\subfigure[Kendall $ \tau $, stride $ \beta $]{
		\label{stride_kendall}
		\includegraphics[width=0.46\linewidth]{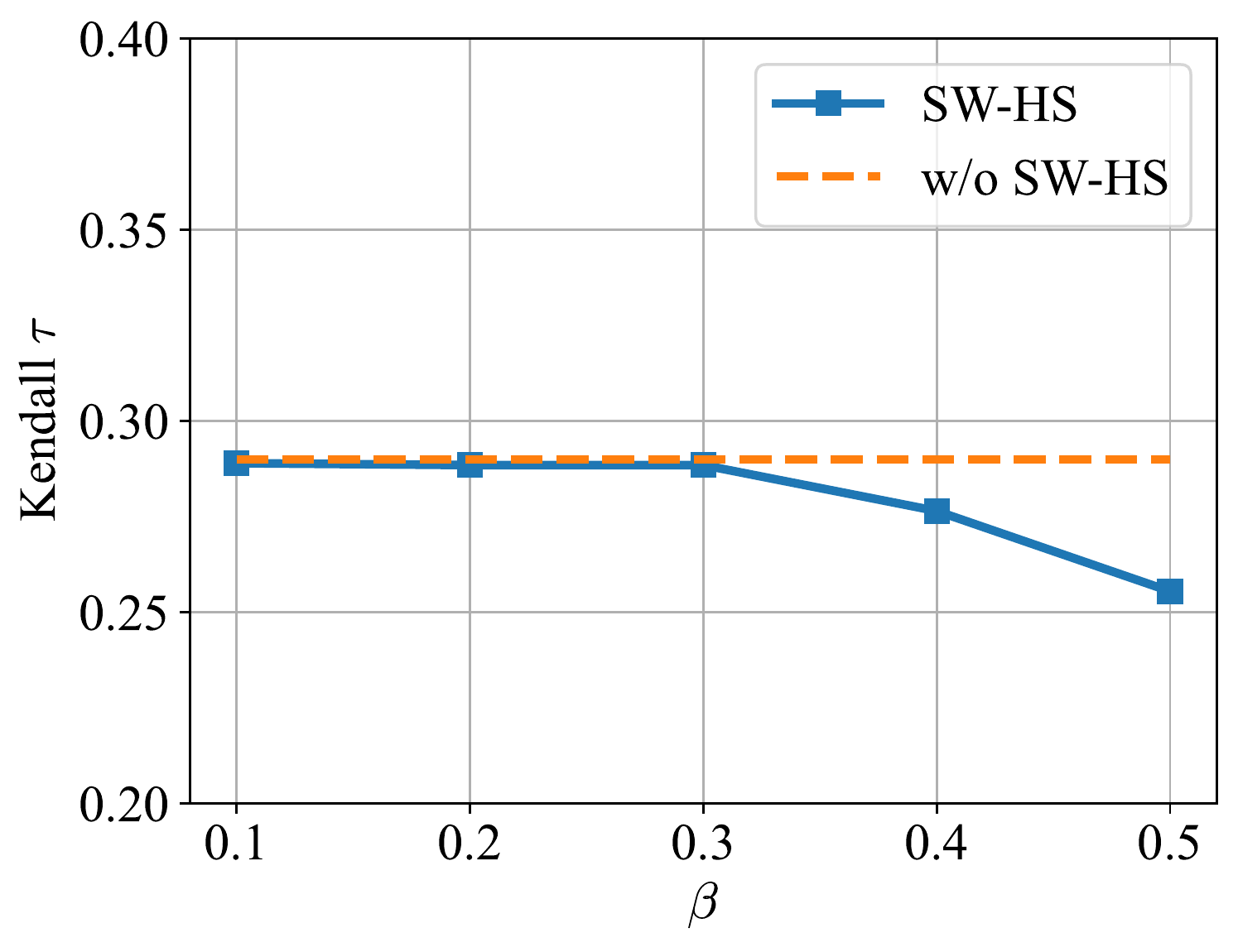}
	}
	\caption{Effects of different configurations of hyper-parameters on Flickr30K.}
\end{figure}
\begin{figure*}[!t]
	\centering
	\includegraphics[width=\linewidth]{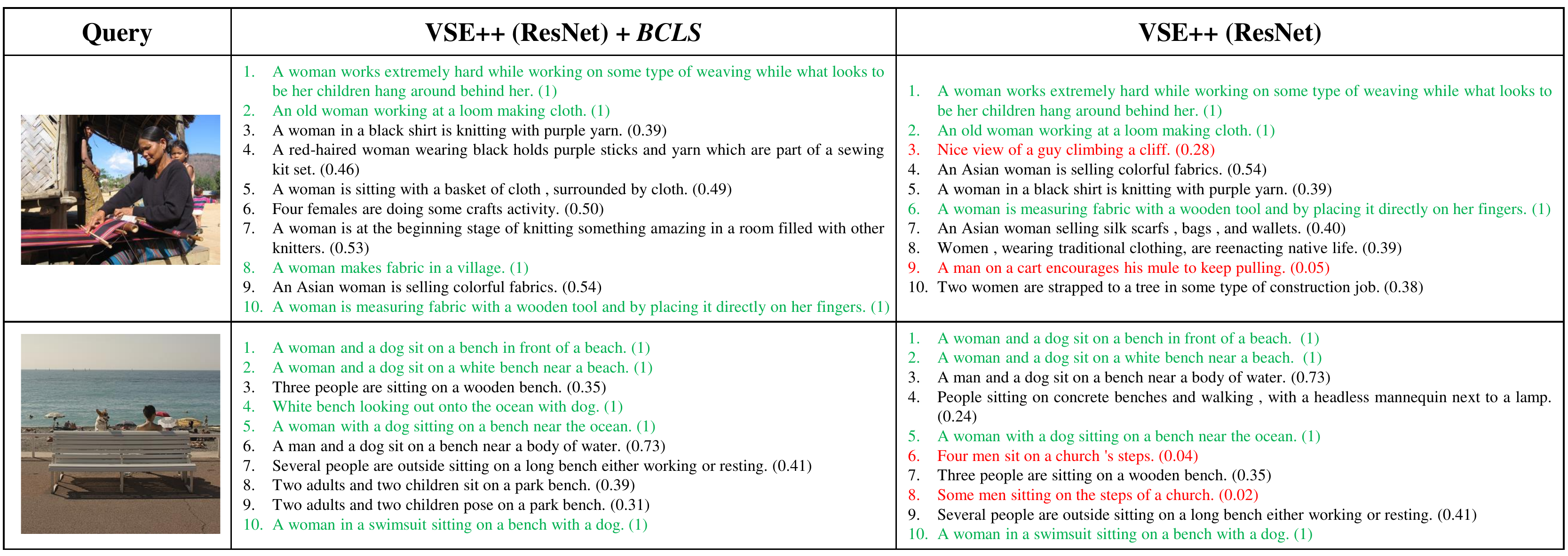}
	\caption{Qualitative image-to-text retrieval comparison between the baseline and our framework with \textit{BCLS} on the Flickr30K test set using VSE++ (ResNet).}
	\label{i2t}
\end{figure*}
\begin{figure*}[!t]
	\centering
	\includegraphics[width=\linewidth]{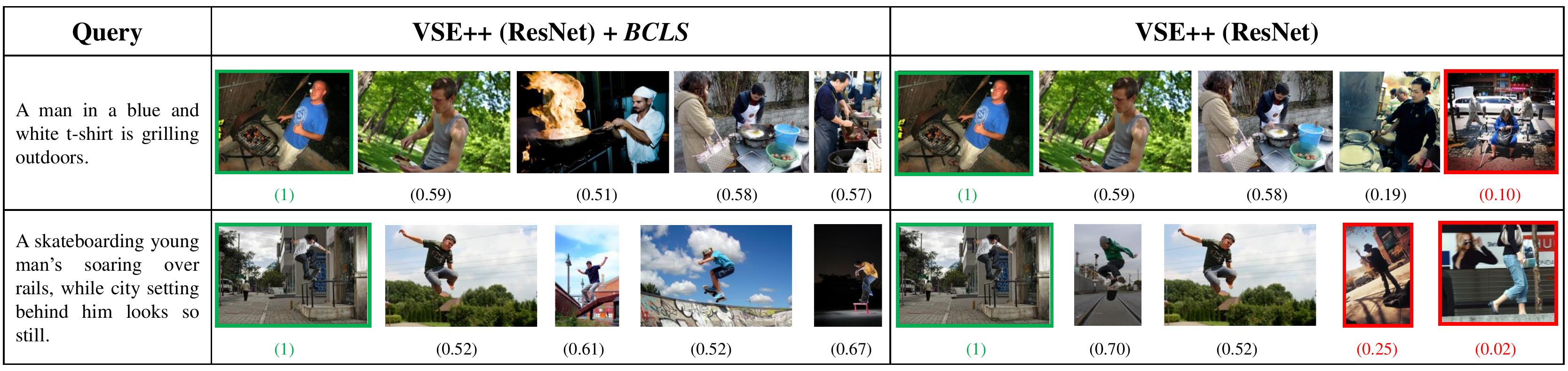}
	\caption{Qualitative text-to-image retrieval comparison between the baseline and our framework with \textit{BCLS} on the Flickr30K test set using VSE++ (ResNet).}
	\label{t2i}
\end{figure*}
In Kendall ranking loss, two parameters relaxation $ \alpha $ and stride $ \beta $ are introduced to control the sampling strategy. 
It is worth exploring a trade-off between retrieval performance and computational complexity. We experiment with several combinations of parameters on Flickr30K using VSE++.

\textbf{Relaxation $ \alpha $:} We test the effect of $ \alpha $ by fixing $ \beta = 0.1 $.
We test RSUM and Kendall $ \tau $ on the test set under different parameter conditions on the Flickr30K dataset. 
The ``w/o BCLS'' in \figurename~\ref{gap_rsum} and \figurename~\ref{gap_kendall} means that only $ \mathcal{L}_{\text{T-SN}} $ is used during training, and $ \mathcal{L}_{\text{K-SW-HS}} $ is not used.
As shown in \figurename~\ref{gap_rsum} and \figurename~\ref{gap_kendall}, when $ \alpha = 0.2 $, both RSUM and \textit{Kendall} $ \tau $ achieve the highest performance. 
According to our statistics, this standard deviation of the textual similarities between multiple captions of the same image on the Flickr30K datasets is around $ 0.2 $. 
This shows that when $ \alpha $ is set to the standard deviation, it can best alleviate the negative impact of inaccurate continuous pseudo labels.

\textbf{Stride $ \beta $:} We test the effect of $ \beta $ by fixing $ \alpha = 0.2 $. 
The ''w/o SW-HS`` in \figurename~\ref{stride_rsum} and \figurename~\ref{stride_kendall} means that the \textit{Sliding Window sampling and Hard Sample mining strategy} (\textit{SW-HS}) are not used during training, which will have a high complexity of $ O(N^{3}) $.
From \figurename~\ref{stride_rsum} and \figurename~\ref{stride_kendall} we can see that the smaller $ \beta $ is, the better the retrieval performance is. 
A smaller $ \beta $ means a higher sampling frequency, which is closer to the definition of \textit{Kendall} $ \tau $. 
Therefore, the choice of $ \beta $ requires a trade-off between retrieval performance and complexity.

\subsection{Qualitative Results} \label{qr}
\figurename~\ref{i2t} and \figurename~\ref{t2i} shows the qualitative comparison between the baseline and our framework with \textit{BCLS} on the Flickr30K test set using VSE++ (ResNet). 
For image-to-text retrieval, given an image query, we show the top-10 retrieved sentences. 
For text-to-image retrieval, given a sentence query, we show the top-5 retrieved images. 
The number in brackets is the pseudo label between the query and candidates that we pre-calculated.
The ground truth retrieval items for each query are outlined in green. 
Obviously, wrong retrieval results are marked in red. 
As shown in \figurename~\ref{i2t} and \figurename~\ref{t2i}, the results retrieved using the framework with \textit{BCLS} are more relevant to the query than VSE++.
Overall, framework with \textit{BCLS} improves the ranking performance of retrieval results and the incorrect retrieval items are semantically closer to the query. 
For example, in the second row of \figurename~\ref{t2i}, the incorrect results retrieved are also all associated with young men skateboarding. 
This benefits from our framework with \textit{BCLS}, which guides the model to learn continuous pseudo labels, so that a coherent embedding space can be learned, where candidates with higher relevance degrees are mapped closer to the query than those with lower relevance degrees. 
On the other hand, there are some obviously wrong results in the retrieval results using only the VSE++ model.
Since VSE++ only learns binary correlations between queries and candidates. 
It only focuses on correctly ranking the ground truth samples and ignores the learning of correlations among other samples.

\section{Conclusion}
In this paper, an image-text retrieval framework with \textit{Binary and Continuous Label Supervision} (\textit{BCLS}) is proposed to guide retrieval models to learn a coherent visual-semantic embedding space.
The framework combines the advantages of binary and continuous labels, and makes targeted improvements for the problems existing in the two types of label learning.
Our proposed method outperforms the baselines by a large margin and obtains competitive results on two image-text retrieval benchmarks.
Moreover, we conduct an objective and fair comparison of existing semantic-based retrieval methods, verifying that our method can better learn continuous semantic relations.
In future work, we plan to extend the framework to other tasks with continuous label supervision.


\bibliographystyle{IEEEtran}
\bibliography{IEEEabrv}

\end{document}